\documentclass[sigconf]{acmart}
\usepackage[utf8]{inputenc}
\usepackage{amsmath,amsthm} 
\usepackage{dsfont}
\usepackage{graphicx}
\usepackage{float}
\usepackage{color}
\usepackage[linesnumbered,lined,boxed,commentsnumbered]{algorithm2e}
\usepackage{subcaption}
\usepackage{enumitem}
\usepackage{xcolor}

\copyrightyear{2022}
\acmYear{2022}
\setcopyright{rightsretained}

\acmConference[KDD '22]{Proceedings of the 28th ACM SIGKDD Conference on Knowledge Discovery and Data Mining}{August 14--18, 2022}{Washington, DC, USA}
\acmBooktitle{Proceedings of the 28th ACM SIGKDD Conference on Knowledge Discovery and Data Mining (KDD '22), August 14--18, 2022, Washington, DC, USA}
\acmISBN{978-1-4503-9385-0/22/08}
\acmDOI{10.1145/3534678.3539211}

\usepackage{etoolbox}
\makeatletter
\patchcmd{\maketitle}{\@copyrightpermission}{
   \begin{minipage}{0.3\columnwidth}
     \href{https://creativecommons.org/licenses/by/4.0/}{\includegraphics[width=0.90\textwidth]{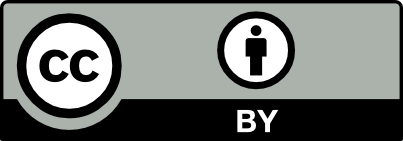}}
   \end{minipage}\hfill
   \begin{minipage}{0.7\columnwidth}
     \href{https://creativecommons.org/licenses/by/4.0/}{This work is licensed under a Creative Commons Attribution International 4.0 License.}
   \end{minipage}
 
   \vspace{5pt}
}{}{}

\makeatother

\begin{document}

\title{
ROI-Constrained Bidding via Curriculum-Guided Bayesian Reinforcement Learning}

\author{Haozhe Wang}
\email{jasper.whz@outlook.com}
\affiliation{
  \institution{ShanghaiTech University}
  \country{China}
}
\author{Chao Du}
\email{duchao0726@gmail.com}
\affiliation{%
  \institution{Alibaba Group}
  \country{China}
}

\author{Panyan Fang}
\author{Shuo Yuan}
\affiliation{%
  \institution{Alibaba Group}
  \country{China}
}
\author{Xuming He}
\affiliation{
  \institution{ShanghaiTech University}
  \country{China}}
\author{Liang Wang}
\affiliation{
  \institution{Alibaba Group}
  \country{China}
}
\author{Bo Zheng}
\affiliation{
  \institution{Alibaba Group}
  \country{China}
}

\renewcommand{\shortauthors}{Haozhe Wang et al.}

\begin{abstract}
    Real-Time Bidding (RTB) is an important mechanism in modern online advertising systems. Advertisers employ bidding strategies in RTB to optimize their advertising effects subject to various financial requirements, especially the return-on-investment (ROI) constraint. ROIs change non-monotonically during the sequential bidding process, and often induce a see-saw effect between constraint satisfaction and objective optimization. While some existing approaches show promising results in static or mildly changing ad markets, they fail to generalize to highly dynamic ad markets with ROI constraints, due to their inability to adaptively balance constraints and objectives amidst non-stationarity and partial observability.
    In this work, we specialize in ROI-Constrained Bidding in non-stationary markets. Based on a Partially Observable Constrained Markov Decision Process, our method exploits an indicator-augmented reward function free of extra trade-off parameters and develops a \emph{Curriculum-Guided Bayesian Reinforcement Learning (CBRL)} framework to adaptively control the constraint-objective trade-off in non-stationary ad markets. Extensive experiments on a large-scale industrial dataset with two problem settings reveal that CBRL generalizes well in both in-distribution and out-of-distribution data regimes, and enjoys superior learning efficiency and  stability.
\end{abstract}



\maketitle

\def\bb{\mathbf{b}}
\def\ecpm{\text{eCPM}}
\def\E{\mathbb{E}}
\def\wr{\text{win\_rate}}
\def\cost{\text{cost}}
\def\rev{d}

\def\S{\mathcal{S}}
\def\A{\mathcal{A}}
\def\P{\mathcal{P}}
\def\O{\mathcal{O}}
\def\EE{\mathcal{E}}
\def\pv{\mathbf{\text{pv}}}
\def\x{\mathbf{x}}
\def\roi{\text{ROI}}
\def\rsum{\text{rsum}}
\def\csum{\text{csum}}
\def\1{\mathds{1}}
\def\I{\mathbb{I}}
\def\RR{\mathbb{R}}
\def\R{\mathcal{R}}
\def\J{\mathcal{J}}
\def\C{\mathcal{C}}
\def\T{\mathcal{T}}
\def\log{\text{log}}
\def\id{\textbf{ID} }
\def\ood{\textbf{OOD} }
\def\sc{\textbf{SC} }
\def\mc{\textbf{MC} }
\def\cbrl{\textbf{CBRL} }
\def\uscb{\textbf{USCB} }
\def\mprice{p_{M|X}(m|\x)}
\def\mpricei{p_{M|X}(b_i|\x)}

\newcommand{\defeq}{\mathrel{\overset{\makebox[0pt]{\mbox{\normalfont\tiny\sffamily def}}}{=}}}

\vspace{-.2cm}
\section{Introduction}
\vspace{-.1cm}
Online advertising~\cite{gsp,ad3} has become an important business in the modern Internet ecosystem, connecting vast amounts of advertisers and users closely. Through Real-Time Bidding (RTB) systems~\cite{ortb}, the online advertising markets manage to process a throughput of billions of ad impression opportunities, each triggering a bidding auction (Fig.~\ref{fig:rtb}). During the online sequential bidding process, the advertisers employ bidding strategies to optimize their advertising effects, subject to the budget constraint, and usually with return-on-investment (ROI) requirements. 
ROI, computed as the ratio of the value obtained to the price paid, is the standard metric to measure the immediate trade-off between (various types of) return and investment. In particular, ROI constraints are widely adopted by \emph{performance advertisers} who concern about the effectiveness of resource used~\cite{performanceAd}. 

In recent years, extensive research has been conducted on constrained bidding. Most of these works focus on the budget-only setting~\cite{knapsack, linear, ortb, RLB, BCB, optsurvey}, and they cannot generalize to deal with the ROI constraints, due to the \emph{non-monotonicity} of ROIs. ROIs can either increase or decrease over time during the sequential bidding process, in contrast to the budget that always decreases. Previous works on budget-constrained bidding derive pacing strategies (c.f. \cite{optsurvey} for a survey) that terminate bidding upon depleted budget (c.f. \cite{BCTR}) or exploit the monotonicity of the budget in a Markov Decision Process (MDP) formulation~\cite{RLB,BCB}. Neither of these approaches complies with the non-monotonic ROI constraints. 

Moreover, ROI-constrained bidding usually witnesses a see-saw effect between constraint satisfaction and objective optimization, which urges the need to balance between constraints and objective. For example, return can increase with ROI decreasing when return and investment grows at different speeds (c.f. Sec.~\ref{sec:ps}). Recent approaches~\cite{MCB, RCPO} handles the constraint-objective trade-off by \emph{soft combination algorithms}, which introduce extra trade-off parameters to softly combine constraint violations and objective value in the objective functions. Despite their promising results, these methods assume static or mildly changing markets, which are limited in the more \emph{non-stationary markets}. Such application scenarios are common when uncontrollable or unpredictable external forces affect the auction markets. For instance, the external online ad markets are prone to unknown adversaries that interfere with auction winning.


With ROI constraints and non-stationarity intertwined, ROI-Constrained Bidding (RCB) in the general ad markets is challenging. On the one hand, the optimal constraint-objective trade-off can vary across different market dynamics. As such, soft combination algorithms that employ a static trade-off parameter design fail to adapt constraint-objective trade-off per dynamics, leading to non-responsive and unidentifiable bidding behaviors. In addition,
the bidders are generally unobservable to other competing bidders in each auction, with market information leaking only conditionally (c.f. Sec.\ref{sec:ps}). Consequently, such \emph{partial observability} makes it even harder for the bidders to coordinate with the market dynamics. 

To address these challenges, we specialize in the problem of ROI-Constrained Bidding (RCB) in non-stationary markets. Based on a Partially Observable Constrained Markov Decision Process (POCMDP) formulation for RCB, we introduce the first hard barrier solution to accommodate non-monotonic constraints(c.f. soft combination solutions). Our method employs the indicator function to render RCB an unconstrained problem, and develops a Curriculum-Guided Bayesian Reinforcement Learning (CBRL) framework to achieve adaptive control of constraint-objective trade-off. 

Specifically, to avoid the pitfalls of soft combination algorithms in non-stationary markets, we introduce a reward function that incorporates the indicator function and involves \emph{no extra trade-off parameters}. The indicator-augmented reward function explicitly encourages feasible solutions over infeasible ones by setting a hard barrier, which removes ambiguity in rewards (Sec.~\ref{sec:indicator}). However, as the reward function inherits the final-time sparsity from ROI that may hinder policy learning due to reward sparsity~\cite{ICM}, we further develop a curriculum learning procedure to address inefficient policy search. By exploiting the problem structure, the curriculum learning arranges a sequence of proxy problems that provides immediate reward signals with an optimality guarantee, leading to faster convergence and better performance (Sec.~\ref{sec:cl}). 

While the parameter-free property of the hard barrier rewards is intriguing, the adaptive constraint-objective trade-off in partially observable markets amidst non-stationarity inherently lends to policy learning. To achieve this, we embrace a Bayesian approach. In particular, the agent learns to express its uncertainty about the market based on its past trajectory, by approximate inference of the posterior~\cite{dvl, prml, blei2017variational}. During deployment, the agent manages to infer the market dynamics, acts towards it, and updates the belief over the market with the latest experience, through an iterative process of posterior sampling~\cite{osband,strens}. As such, the agent turns out a Bayes-optimal bidder that achieves the exploration-exploitation trade-off in unknown environments, meanwhile balancing the constraint-objective trade-off (Sec.~\ref{sec:bl}).

We evaluate the proposed CBRL framework on a large-scale industrial dataset, including two problem settings for different advertisers. Extensive experiments verify our superiority over prior methods in both constraint satisfaction and objective maximization, and demonstrate several favorable properties regarding stability and out-of-distribution generalization. Our contributions are:
\begin{itemize}[leftmargin=*, nolistsep]
    \item We present the first hard barrier solution to deal with non-monotonic constraints, which achieves adaptive control of the constraint-objective trade-off in non-stationary advertising markets, and empirically found to reach a new state-of-the-art.
    
    \item We develop a novel curriculum-guided policy search process that promotes efficient policy learning against reward sparsity. 
    
    \item We propose a Bayesian approach that learns adaptive bidding strategies in the partially observable non-stationary markets.

\end{itemize}

\vspace{-.2cm}
\section{Problem Statement}\label{sec:ps}
\vspace{-.1cm}
\newcommand{\rtbplot}{
    \begin{figure}[t]
        \centering
        \includegraphics[width=\linewidth]{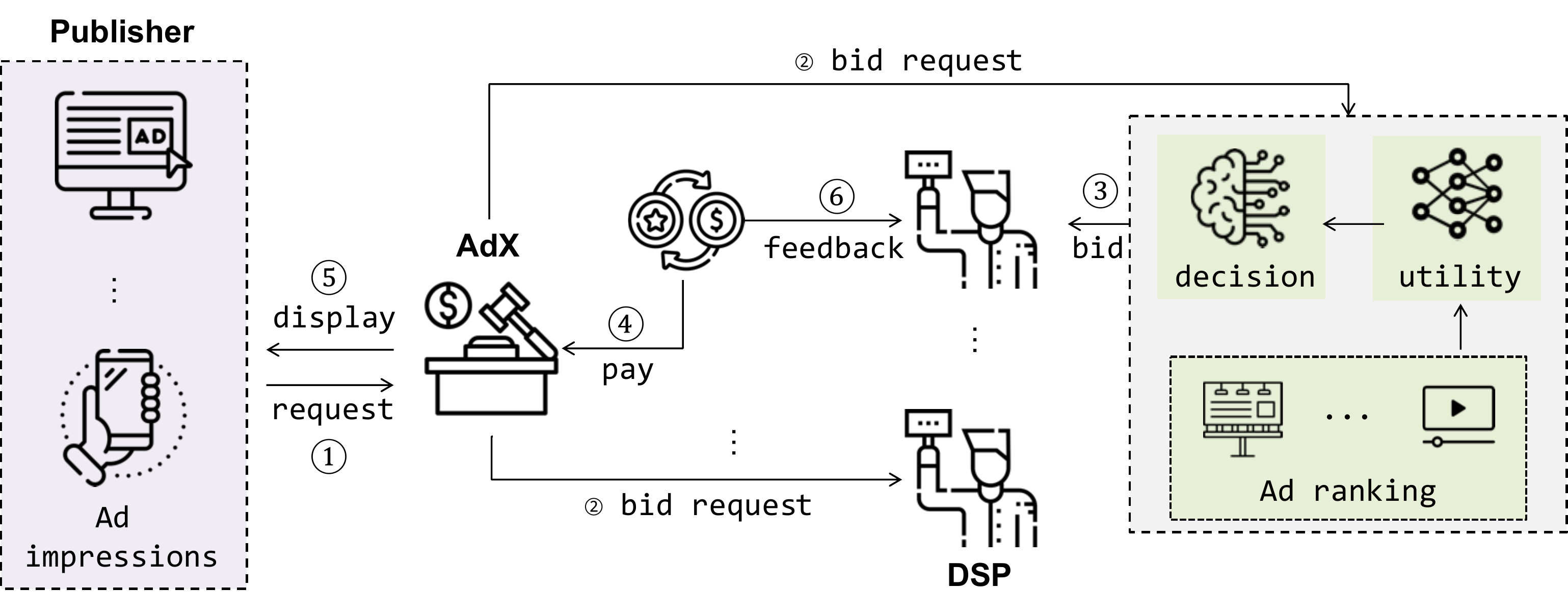}
        \vspace{-.3cm}
        \caption{\small \textbf{An RTB auction.} Ad Exchanger (AdX) broadcasts the ad request to the advertisers. The ad system make decisions and sends the bid. AdX sends win notice, gets paid and diplay the ad. The winner gets delayed feedbacks.}\label{fig:rtb}
        \vspace{-.3cm}
    \end{figure}
}

\rtbplot
Real-Time Bidding (RTB) has become a prevailing advertising paradigm that allows the advertisers to make decisions for every ad impression opportunity~\cite{ortb,ad3}. In RTB, billions of ad impression opportunities arrive sequentially, each triggering an auction. To relieve the advertisers' burden, demand-side platforms (DSPs) offer programmatic buying agents to the advertisers. On behalf of the advertisers, these bidding agents bid for each impression based on the ad context, in an attempt to optimize the hindsight total delivery under financial constraints.

Fig.~\ref{fig:rtb} shows the workflow of each bidding auction. An ad exchanger (AdX) receives a bid request from the publisher when a user triggers an ad impression. AdX then holds an auction and broadcasts the request to all participating DSPs. The bidding agent of each DSP achieves real-time bidding by a modularized bidding engine. The bidding engine first selects an ad targeting the user based on superior personalization techniques, e.g., utility estimations and ad ranking, and decides the bid based on the valuations. Receiving all the bids, AdX announces the highest-bid bidder as the winner, and sends the winner's ad to the publisher for display. The winner pays a charge for the impression opportunity according to the regulated pricing mechanism (e.g., Generalized Second Price~\cite{gsp}), and receives delayed feedback from the publisher. 

Auctions as such take place repeatedly within a period, forming a sequential bidding process for which the advertisers expect to optimize some desired delivery subject to certain constraints. Since RTB is a complex problem that additionally involves personalization techniques~\cite{rank1, du2021exploration} and auction mechanism design~\cite{gsp}, in this work we focus only on the constrained bidding problem, and assume that utility estimations and mechanism design are given beforehand.

Particularly, in this work, we discuss a class of constrained bidding problems, \emph{ROI-Constrained Bidding (RCB)}, which is a major concern of various advertisers. Formally, suppose a bidder observes a bid request $\x_i$ (features about the impression context, and the selected ad) for impression $i$. A bid $b_i$ is decided based on the bidder's estimated utility $u_i$ for the impression. If the bid $b_i$ is larger than the competing market price $m_i$ (i.e., the highest bid of the competing bidders), the bidder wins, pays a cost $c_i$, and  receives delayed feedback about the delivery $d_i$ ($u_i$ estimates $d_i$). The RCB problem aims to maximize the total delivery subject to a budget $B$ and a \emph{return-on-investment (ROI)} constraint limit $L$, within $T$ impressions. 
\begin{equation}
     \begin{aligned}
    \max_{\mathbf{b}} \quad
    &D(\epsilon_T),\quad
    \text{s.t} \quad \roi(\epsilon_T) \geq L, \quad
    B-C(\epsilon_T)\ge 0
\end{aligned}\label{obj}
\end{equation}
where we denote $\epsilon_t = \{(b_i,m_i,c_i, u_i,d_i)\}_{i=1}^t$ as a $t-$step episode containing $t$ impressions, and we introduce the following notations to denote the cumulative delivery, cost, and ROI of an episode $\epsilon_t$,
\begin{equation}
    \begin{aligned}
    D(\epsilon_t)\defeq \sum_{i=1}^t d_i~\1_{b_i>m_i},
    C(\epsilon_t)\defeq\sum_{i=1}^t c_i~\1_{b_i>m_i},
    \roi(\epsilon_t) \defeq \frac{D(\epsilon_t)}{C(\epsilon_t)}
    \end{aligned}\label{roidef}
\end{equation}
and use the short-hands $D(\epsilon_t)\equiv D_t$, $C(\epsilon_t)\equiv C_t$, and $\roi(\epsilon_t)\equiv \roi_t$ if no misconception may arise. 

It is noteworthy that, many widely adopted cost-related key performance indicator (KPI) constraints are viewed as a type of ROI constraint. For example, upper-bounded cost per acquisition (CPA) is equivalent to a lower-bounded per-cost acquisition in the context of ROI constraints. Besides, in this work, we treat \emph{delivery} as equal to \emph{return} for simplicity, which is often the case but exceptions do exist, e.g., profit maximization with per-cost income constraints\footnote{We discuss a generalized version of RCB in our recent work.}.

\noindent \textbf{Challenges of RCB and Related Work.} While the constrained optimization problem~\eqref{obj} appears simply a programming problem, real-world RCB is challenging due to the properties of ROIs, the properties of online ad markets, and both properties intertwined.

Eq.~\eqref{roidef} shows that ROI can either increase or decrease during the bidding process, since both $D_t$ and $C_t$ increase at an uncertain rate. Besides non-monotonicity, a see-saw effect often emerges between constraint violations and the delivery value, esp. when the delivery grows with the investment at a different speed. For example, revenue increases as the cost grow, but the per-cost revenue (ROI) may plunge, inducing the demands of constraint-objective trade-off.

Most existing works on constrained bidding focus on the budget-only setting (c.f. \cite{optsurvey} for a survey), they cannot generalize to deal with ROI constraints. Based on the primal-dual framework, many works derive pacing strategies that terminate bidding when the budget runs out (c.f. \cite{BCTR}). Alternatively, Reinforcement Learning (RL) formulations have been proposed~\cite{RLB,BCB}, which encode the budget status in action space or state space.  

In contrast to these work that exploit the monotonicty of budget, some works propose to deal with specific non-monotonic constraints~\cite{PID,BCTR,RM} or general constraints~\cite{RCPO,MCB}. Among them, a promising solution~\cite{RCPO,MCB} adopts a \emph{soft combination} design that softly combines the constraint violations and the delivery value in the objective function with extra trade-off parameters, theoretically grounded by Lagrangian relaxation to achieve a balanced constraint-objective trade-off. These works, however, are typically established in controlled markets, where market dynamics change smoothly as each ad campaign binds to similar types of impressions, and full access to market information can be gained\footnote{When the publisher offers programmatic advertising services, e.g., in-station advertising of e-Commerce platforms, market information of all bidders is logged.}. 

By contrast, the external online advertising markets experience more drastic market changes, due to unexpected adversaries and system failures. In addition, partial observability of the market aggravates the difficulty of bidding amidst non-stationarity. The bidders can be observable to the market conditionally, or completely unobservable. In particular, under the second-price auctions~\cite{gsp}, the market price $m_i$ equals the cost $c_i$ when the auction is won ($b_i>m_i$), which leaks hindsight information of the market.

In such partially observable markets amidst non-stationarity, adaptive control of the constraint-objective trade-off is hard to achieve, as the optimal trade-off varies across dynamics which is in turn unidentifiable. Previous soft combination solutions rely on static trade-off parameters, logically and empirically found to fail in non-stationary advertising markets (Sec.~\ref{sec:sota}). To this end, in this work, we shed light on an alternative hard barrier solution to accommodate non-monotonic constraints, which learns adaptive bidding strategies per dynamics.

\vspace{-.2cm}
\section{Method}
\vspace{-.1cm}
\newcommand{\modelplot}{
    \begin{figure}[t]
        \centering
    \includegraphics[width=\linewidth]{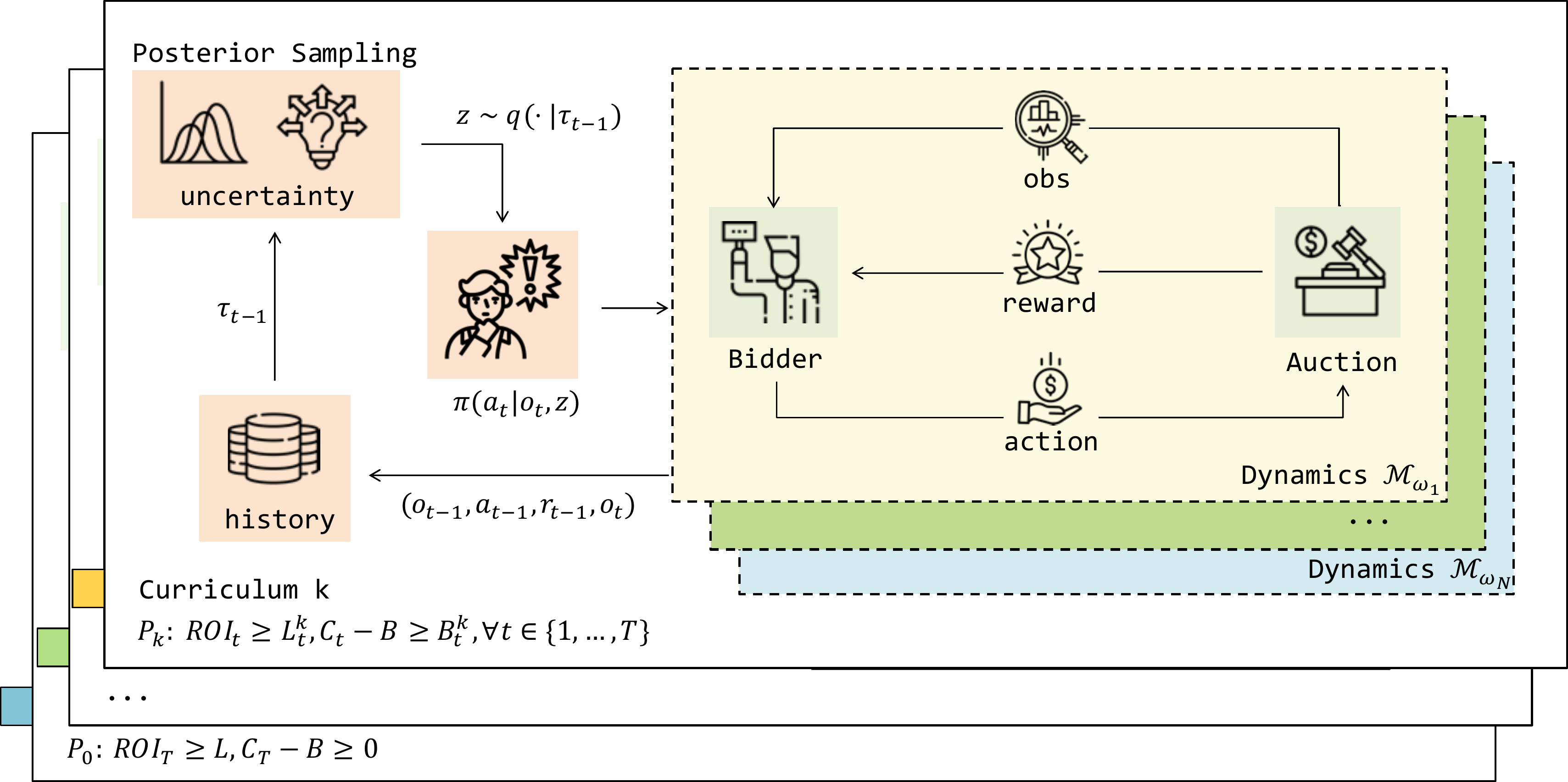}
    \vspace{-.2cm}
    \caption{\small{ \textbf{Model Overview.} We propose CBRL for RCB. Guided by a curriculum sequence, the agent learns to adaptively trade-off constraints and the objective in non-stationary markets. During deployment, the agent updates its belief over the market dynamics based on its past experience, and acts through posterior sampling.}}\label{fig:model}
    \vspace{-.3cm}
    \end{figure}
}
\subsection{MDP Formulation}\label{sec:pomdp}
Markov Decision Processes (MDPs) provide a mathematical framework widely used to learn from interaction with the environment~\cite{suttonRL}. To account for the constraints and the partial observability typical of RCB, we formulate a \emph{Partially Observable Constrained MDP (POCMDP)}, which is finite with $T$ steps, represented as a nine-tuple $\mathcal{M}=(\S,\A,\O, \EE, \T, \mu,\gamma, \R, \C)$:
\begin{itemize}[leftmargin=*]
    \item $\S$. The state space reflects critical information for each impression. Each state $s_i\in \S$ includes impression-level information $(\x_i, d_i, c_i, m_i, u_i)$ and the cumulative statistics $(D_{i-1}, C_{i-1}, \roi_{i-1}, B, L)$\footnote{These statistics cumulates up to the last impression $i-1$ because the feedback of current impression $i$ are received only after the bid.}.
    
    \item $\A$. The action $a_i\in\A$ is a real-valued scalar $b_i\in \RR^+$.
     
    \item $\O$ and $\EE$. The observation space accounts for the partial observability. The emission function $\EE: \S\mapsto \O$ maps a state $s_i$ to an observation $o_i$ by removing $(d_i,c_i,m_i)$.
    
    \item $\T$ and $\mu$. Both symbols determine the market dynamics. The transition probability density $\T(s_{i+1}|s_i,a_i)$: (1) explains the transitions between the cumulative statistics, e.g., $D_{i} = D_{i-1} + d_{i}\1_{b_{i}>m_{i}}$, which are markovian; and (2) induces stochasity from the market dynamics, i.e., $P(\x|i),P(d,c,m|\x_i)$, which are time-varying. The initial state distribution $\mu(s_1)$ can be arbitrary. While we have no access to the exact form of $\T,\mu$, we approximate these with the empirical distribution using logged dataset.
    \item $\gamma$. The discount factor. 
    \item $\R$ and $\C$. We define the following reward function and cost function to account for the performance objective and constraint requirements respectively. 
    \begin{align}
        &\mathcal{R}(s_i,a_i) = \left(D_{i}-D^{-}\right)~\1_{i=T}, \label{rsparse}\\
        &\mathcal{C}(s_i,a_i) = \left((L-\roi_i)~\1_{\bar{F}_L}+(C_{i}-B)~\1_{\bar{F}_B}\right)\cdot\1_{i=T}\label{csparse}
    \end{align}
    We note the above functions are piece-wise functions that only evaluate at termination. We define $D^{-}\defeq \inf D_T$ so that $\R(s_T,a_T)>0$. To simplify the notation of feasibility, we use $F_L(\epsilon_t)\defeq \{\epsilon_t~|~\roi(\epsilon_t)\ge L\}, F_B(\epsilon_t)\defeq \{\epsilon_t~|~C(\epsilon_t)\le B\}, F(\epsilon_t)\defeq F_L(\epsilon_t) \cap F_B(\epsilon_t)$ to indicate the feasible solution sets that respect the ROI constraint, the budget constraint and both constraints. By convention, $\bar{F}_L,\bar{F}_B,\bar{F}$ are their negations.
\end{itemize}

\noindent The RL objective for the above MDP is:
\begin{equation}
\begin{aligned}
\underset{\pi}{\max}~\E_{} \left[
    \sum\nolimits_{t=1}^T \mathcal{R}(s_t,a_t) 
    \right], 
~\text{s.t.}~\E_{} \left[
    \sum\nolimits_{t=1}^T \mathcal{C}(s_t,a_t) 
    \right] \le 0
\end{aligned}\label{rlobj2}
\end{equation}

We remark that both the objective value and the constraints are studied in expectation. The expectation is taken over different problem instances, which coincides with the fact that real-world advertisers run various ad campaigns or consider advertising effects over different time periods. As we encode the stochasticity and variations of the market dynamics in $(\T,\mu)$, the RL objective aligns with RCB on an aggregate level. Moreover, cost function~\eqref{csparse} is a subtle yet noteworthy design that doesn't violate constraint satisfaction in the expected sense\footnote{Non-negative entries less than $0$ in expectation means each entry must evaluate $0$.}. 

\subsection{Curriculum-Guided Bayesian Reinforcement Learning}
In this section, we present a \emph{Curriculum-Guided Bayesian Reinforcement Learning (CBRL)} framework (Fig.~\ref{fig:model}) to solve the MDP. Specifically, we tackle the long sequence with a slot-wise policy design (Sec.~\ref{sec:semi}), accommodate the constraints with a parameter-free hard barrier reward function (Sec.~\ref{sec:indicator}), promote efficient policy learning by curriculum-guided policy search (Sec.~\ref{sec:cl}) and achieves adaptive control of constraint-objective trade-off in non-stationary markets with Bayesian RL (Sec.~\ref{sec:bl}). 

\subsubsection{A semi impression-level policy to deal with long sequences.}\label{sec:semi}
While bidding agents aim to bid in real-time for each impression, real-world advertising systems experience a throughput of billions of impressions. This brings about extremely long trajectories for RL that incurs training issues. The optimal bidding theorem comes to the rescue, providing a proxy for converting impression-level bidding to a ratio controlling problem.

\begin{theorem}
    In second-price auctions, the optimal bidding function for problem~\eqref{obj} is:
    \par\vspace{-.2cm}
    \begin{equation}
        b_i = \beta~u_i, \quad \beta>0
    \end{equation}\label{optb}\label{thm1}
    \par\vspace{-.4cm}
\end{theorem}

The theorem (proof in the appendix) states that the optimal bid for each impression is linear w.r.t. the impression value $u_i$, governed by a bid ratio $\beta$ computed given the problem data. Intuitively, the bid ratio $\beta$ acts like a threshold in the knapsack problem where items are filtered by their value-to-weight ratios~\cite{optsurvey}. 

Based on Thm.~\ref{thm1}, we reduce the online RCB problem into a ratio controlling problem in which we seek to find optimal bid ratios for binned time slots. The reduction makes policy trajectories tractable to train with, though at the cost of precisely impression-level control. 
Specifically, we set the slot length to the unit time period of dynamics changes (empirically thirty minutes). Indexing each time slot by $t$, we add an auxiliary action space $a_t\equiv \beta_t \in \A^\prime$ for the slot-wise bid ratios, and an observation space $o_t\in \O^\prime$, which replaces the impression-level information with aggregate-level statistics, e.g., the slot-wise delivery $D_t-D_{t-1}$ (more details in the appendix). We note that this slot-wise policy can still produce impression-level bids with the impression-level utilties:
\begin{equation}
    b_i = \beta_t~u_i,\quad \text{where}\quad \beta_t  \sim \pi(\cdot| o_t)
\end{equation}

\subsubsection{An indicator-augmented reward function to accommodate constraints.}\label{sec:indicator}
Our goal is to develop a parameter-free hard barrier solution to address problem~\eqref{obj}. The main idea is to convert the constrained optimization problem to an unconstrained counterpart, which derives a parameter-free reward function that still satisfies Bellman Equation to fit in with conventional policy optimization. 

Specifically, we note that the RL objective~\eqref{rlobj2} has an equivalent unconstrained form $\max_\pi \E \left[\sum_{t=1}^T\R^\prime(s_t,a_t)\right]$, with an indicator-augmented reward function defined as follows,
\begin{equation}
    \begin{aligned}
        \R^\prime(s_t,a_t)=\left(
        \1_{F}\sum\nolimits_{t=1}^T \mathcal{R}(s_t,a_t) 
        - \1_{\bar{F}}\sum\nolimits_{t=1}^T \mathcal{C}(s_t,a_t)\right)\cdot \1_{t=T}.\label{rprime}
    \end{aligned}
\end{equation}

The equivalence holds because $\sum\nolimits_{t=1}^T \mathcal{R}(s_t,a_t)> 0\ge-\sum\nolimits_{t=1}^T \mathcal{C}(s_t,a_t)$ strictly holds. Another critical property of Eq.~\eqref{rprime} is it satisfies the recursive property of Bellman Equation and thus works with conventional RL algorithms.

While this reward function design appears simple, it renders a parameter-free solution to accommodate constraints, by setting a hard barrier between feasible and infeasible solutions. The underlying philosophy is that, we designate feasibility to weigh heavier than infeasibility in rewards, instead of softly combining the constraint violations and delivery value as adopted in soft combination algorithms~\cite{MCB,RCPO}. Soft combination solutions are ambiguous in reflecting the quality of different solutions, especially when the trade-off parameters are inappropriate. For example, an infeasible solution with inadequate penalties for constraint violations would be favored over a feasible solution. Another drawback of soft combination algorithms is that static trade-off parameters may be inapt for dynamic markets (Sec.~\ref{sec:sota}).


\subsubsection{Curriculum-guided policy search to promote efficient learning.}\label{sec:cl}
While the reward function~\eqref{rprime} enjoys several favorable properties, we note that its reward sparsity hinders efficient policy learning. In particular, the agent is only guided with reward signals at termination, leading to inefficient and blinded exploration in the policy search space~\cite{ICM}. To resolve this, our goal is to provide the agent with immediate rewards to guide policy learning, while preserving the optimality of convergence. With proxy problems to constrained problem~\eqref{obj} that provide dense signals, we arrange a sequence of proxies into a curriculum learning procedure that regularizes policy search and guides the policy towards optimality. 


We begin by defining the following problem $\textit{P}_k (k>0)$ as a proxy to problem~\eqref{obj}, denoted as $\textit{P}_0$.
\begin{equation}
    \begin{aligned}
   \max_{\mathbf{b}} 
   D_T~
   \quad\text{s.t.} 
   \quad \roi_{t} \ge L_t^k, 
   B-C_{t}\ge B_t^k,
   ~\forall t\in \{1,\dots,T\},
\end{aligned}\label{proxy}
\end{equation}
where we add $T-1$ constraints in each time slot, except for $T-$th slot where $L_T^k=L,B_T^k=0$. These additional constraints admit the potential for immediate rewards. 


To arrange a meaningful curriculum sequence which starts with easier problems and converges to the original problem~\eqref{obj}, we consider two questions. How do we exploit the dense constraints to provide dense signals? And how do we arrange the curricula?

For the first question, we adopt a greedy approximation that derives a dense reward function. Treat $\textit{P}_k (k>0)$ as a recursive stack of sub-problems in each time slot $t$, we define recursively for each time slot:
\begin{equation}
    \begin{aligned}
   &\max_{\mathbf{b}} 
    \quad D_t=D^\ast_{t-1}+D_{t-1:t}~\\
   &\text{s.t.} 
   \quad\roi_{t} \ge L_t^k, 
   \quad B-C_{t}\ge B_t^k,
   \forall t\in \{1,\dots,T\}.
\end{aligned}\label{proxy2}
\end{equation}
The objective of the above recursive sub-problem is to greedily maximize $D_{t-1:t}$ under the added slot-wise constraints, based on $D^\ast_{t-1}$ obtained in the previous slot. As a result, this recursive structure offers a \emph{dense} reward function:
\begin{equation}
    \begin{aligned}
        \R^{\prime}_k(s_t,a_t) = D_{t-1:t}\1_{F} -\left(L_t^k-\roi_t\right) \1_
        {F_{L_t^k}}
        -\left(B_t^k+C_{t}-B\right)\1_
        {F_{B_t^k}},
    \end{aligned}
     \label{rdense}
\end{equation}
which credits the agent with the slot-wise delivery if cumulative constraints are satisfied, and penalizes the agent with constraint violations if otherwise. 

To arrange the curricula into meaningful sequence, the idea is to evolve the constraints of the proxies from tightened ones to loosened ones, until approaching $\textit{P}_0$. As we deploy each curriculum as a dense reward function, the curriculum sequence $\{\textit{P}_n,\dots,\textit{P}_k,\dots, \textit{P}_0\} (n>k)$ is implemented as a dense reward function with constraint limits $\{L_t^k\}_{t=1,k=1}^{T,n},\{B_t^k\}_{t=1,k=1}^{T,n}$ evolving according to the following principles: (1) The limits increase as $k$ increases, so tighter constraints are levied on earlier curricula; (2) $\lim_{t\rightarrow T} L_t^k = L$ and $\lim_{t\rightarrow T}B_t^k = 0$, so that the constraint limits approach the final constraint as time goes. The specific design of the constraint limits can be found in the appendix. 

Curriculum learning of this kind starts from proxy problem $\textit{P}_n$, with tightened constraints that strongly narrow the policy search space. However, these strong signals might bias toward sub-optimal behaviors, so we proceed with problems with loosened constraints, and finally, approach problem $\textit{P}_0$. The curriculum-guided policy search constructs recursive sub-problems of RCB, and promotes faster convergence by dense reward signals than direct optimization with sparse rewards. Although logically curriculum learning requires multiple stages of training, policy learns efficiently in each curriculum so that the overall training time is less than policy learning with sparse rewards (Sec.~\ref{sec:expcl}). To relieve hand-tuning of the curriculum settings, we develop an automated curriculum learning process driven by differentiable regret minimization, as detailed in the appendix. 

\modelplot
\subsubsection{Bayesian reinforcement learning to act optimally amidst non-stationarity and partial observability.}\label{sec:bl}
In uncontrollable markets, the bidder has no access to the complete auction market information (i.e., $(d_i,c_i,m_i)$), and the market dynamically changes due to unknown time-varying parameters (i.e., $P(d,c,m|\omega_t)$). We summarize these factors as partial observability in the POCMDP formulation. 

Adapting the constraint-objective trade-off per dynamics is challenging. To resolve this, we adopt a Bayesian perspective~\cite{BAMDP, BRL}. In tabular POMDPs, policies that perform posterior sampling~\cite{osband,strens} given the belief over MDPs have proven Bayes-optimal~\cite{BAMDP,BAMDP2}, meaning to balance exploration and exploitation in an unknown environment. Motivated by this, we aim to first infer the posterior of the unobservable market dynamics via Variational Bayes~\cite{dvl} and then act adaptively through posterior sampling. 

Specifically, we adopt a latent representation $z$ for the unobservability. Our goal is to learn a variational distribution $q(z|\tau_{t})$ that allows the agent to approximately infer the posterior $P(z)$, based on the historic trajectory $\tau_{t}=\{(o_i,a_i,o_{i+1})\}_{i=1}^t$. In Variational Bayes, the variational distribution is typically learned with an \emph{Evidence Lower Bound (ELBO)}~\cite{dvl}. Recall that the Q-learning objective is:
\begin{equation}
    \E_{\mathbf{c}_t\sim\mathcal{B}}\left[ \left(Q(o_t, a_t)-\left(r_t+\gamma \max_a Q(o_{t+1},a)\right)\right)^2\right]\label{qlearning}
\end{equation}
where $\mathcal{B}$ denotes a replay buffer~\cite{dqn}, and $\mathbf{c}_t\defeq (o_t,a_t, o_{t+1}, r_t)$.

Minimizing Bellman Residual can be interpreted as maximizing the log likelihood of the transition tuple $(o_t,a_t, o_{t+1}, r_t)$ in a model-free manner~\cite{caster}, and thus we maximize the ELBO of Eq.~\eqref{qlearning} to learn the inference network $q(z|\tau_{t})$ (derivations in the appendix). 
\begin{equation}
\begin{aligned}
    \E\left[
        -\E_{z}\left[ 
    \left(Q(o_t, z, a_t)-y_t\right)^2\right]
    - D_{KL}\left(q(\cdot|\tau_{t-1})|P(z)\right)\right]
\end{aligned}\label{elbo}
\end{equation}
where $z\sim q(\cdot|\tau_{t-1})$ and $\mathbf{c}_t, \tau_{t-1}\sim\mathcal{B} $ is omitted, the target value\footnote{The target value can be computed by other formulas according to the algorithm used.} computes as $y_t\defeq r_t+\gamma \max_a Q(o_{t+1},z_t,a)$. $P(z)$ is set to $\mathcal{N}\left(0,1\right)$.

The learned Gaussian distribution $q(z|\tau_t)$ expresses our uncertainty about the current dynamics based on our experience, and is implemented as a neural network. Since $q(\cdot|\tau_t)$ is iteratively updated along the trajectory $\tau_t$ up to step $t$, its specific network structure should input variable number of transition tuples $\{\mathbf{c}_i\}_{i=1}^t$ and effectively exploit the relationships between the tuples $\{\mathbf{c}_i\}_{i=1}^t$ that are inherently correlated~\cite{caster}. We adopt the transformer blocks~\cite{vaswani2017attention} followed with average pooling, 
\begin{equation}
    q(z|\tau_{t})=\mathcal{N}\left(\text{AvgPool}\left(f^\mu(\tau_t)\right), \text{AvgPool}\left(f^\sigma(\tau_t)\right)\right)
\end{equation}
where $f^\mu,f^\sigma$ represents a three-layer transformer for the mean and standard deviation distribution parameter.


Equipped with the approximate posterior, the agent performs posterior sampling~\cite{strens, osband} to act in the unknown environment during deployment.
Posterior sampling begins with a hypothesis $z_{t-1}$ sampled from $q(z|\tau_{t-1})$, based on past experience $\tau_{t-1}$. The agent $\pi(a_t|z_{t-1}, o_t)$ acts towards the hypothesized MDP characterized as $z_t$, and collects a new transition tuple $\mathbf{c}_t$. The tuple is used to update the belief $q(z|\tau_{t})$, which reflects the current uncertainty of the market given the agent's past experience. Intuitively, this iterative process allows the agent to test its hypothesis in a temporally extended exploration process, leading to Bayes-optimal exploration-exploitation trade-off in the POCMDP~\cite{BAMDP,BAMDP2}. 

Recall that the reward function~\eqref{rprime} induces a parameter-free form that inherently lends the commonly explicit constraint-objective trade-off to policy learning. The Bayesian bidder complements the reward function by learning to trade-off according to the market dynamics. To this end, the proposed hard barrier solution bears no extra parameters for hand-tuning (c.f. USCB~\cite{MCB}), and performs adaptive control in non-stationary ad markets even with out-of-distribution (OOD) data (Sec.~\ref{sec:expbl}). The insight is that our agent learns to infer the posterior of market transitions in the in-distribution data, which overlaps with that of the OOD data, despite the data distribution being different.

\vspace{-.2cm}
\section{Experiments}
\vspace{-.1cm}
\newcommand{\SotaPlot}{
\begin{figure*}[ht]
    \vspace{-.3cm}
    \centering
    \begin{subfigure}[b]{0.3\textwidth}
         \centering
         \includegraphics[width=\textwidth]{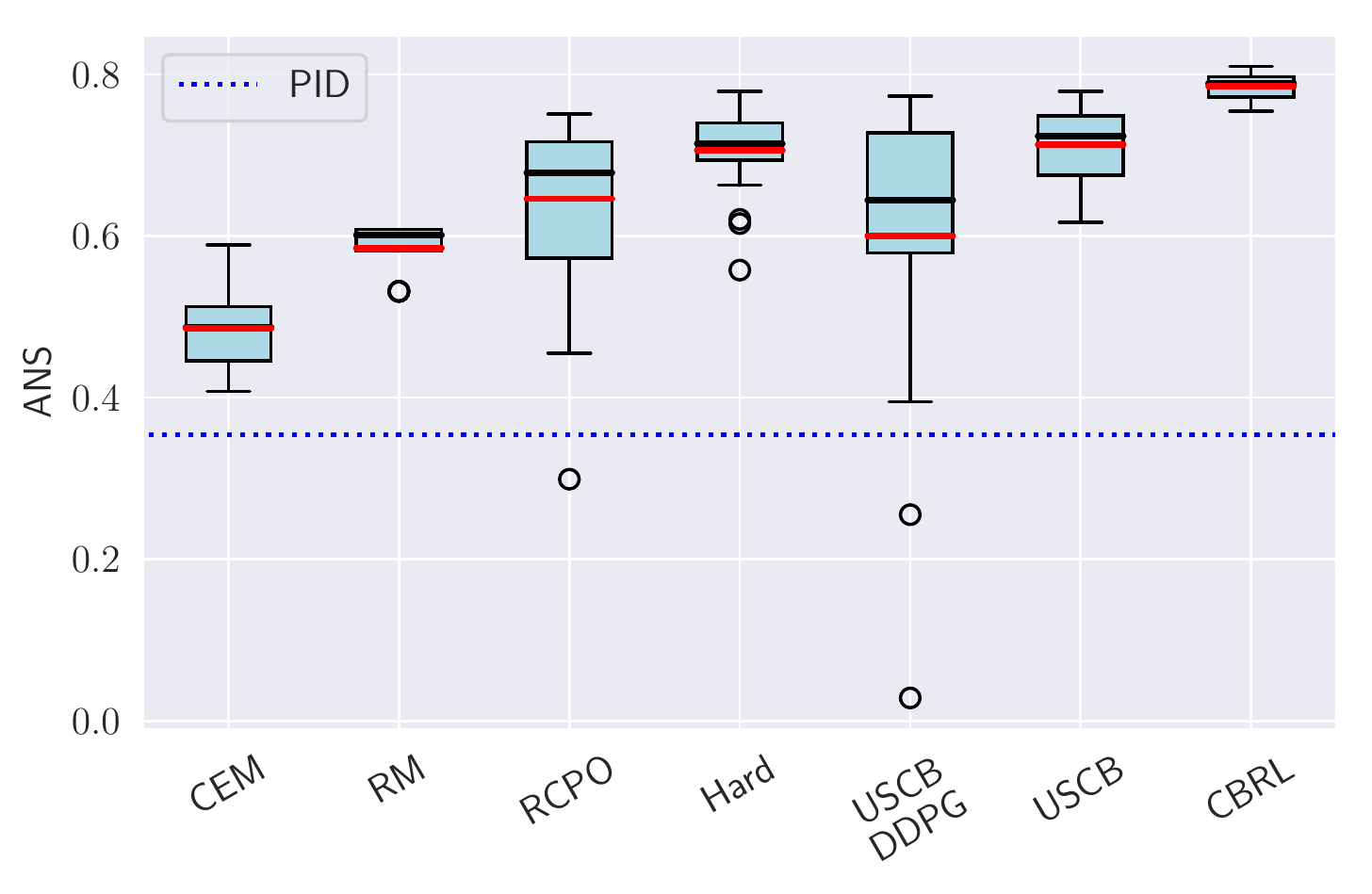}
     \end{subfigure}
     \hfill
     \begin{subfigure}[b]{0.3\textwidth}
        \centering
        \includegraphics[width=\textwidth]{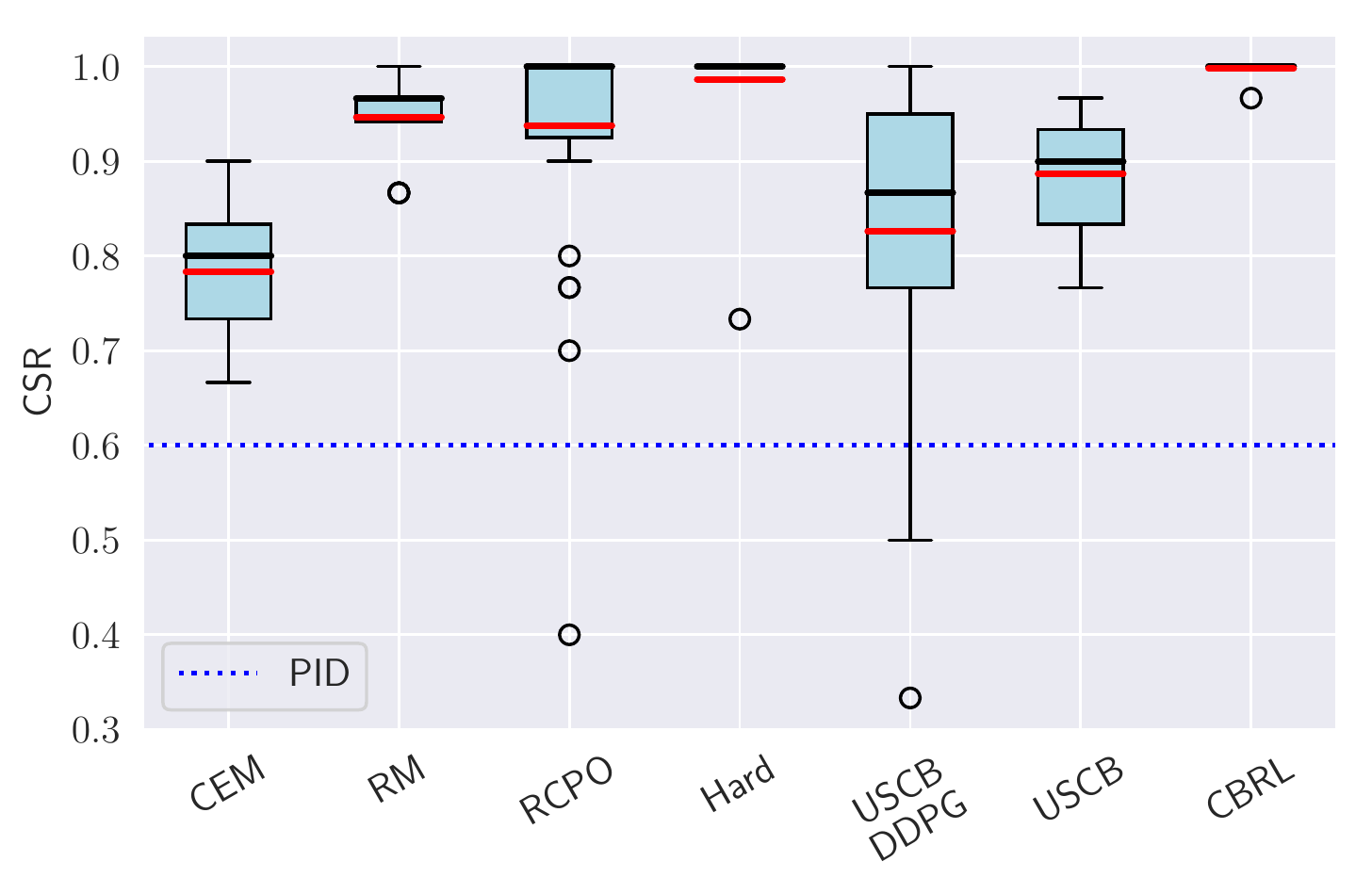}
    \end{subfigure}
    \hfill 
    \begin{subfigure}[b]{0.3\textwidth}
        \centering
        \includegraphics[width=\textwidth]{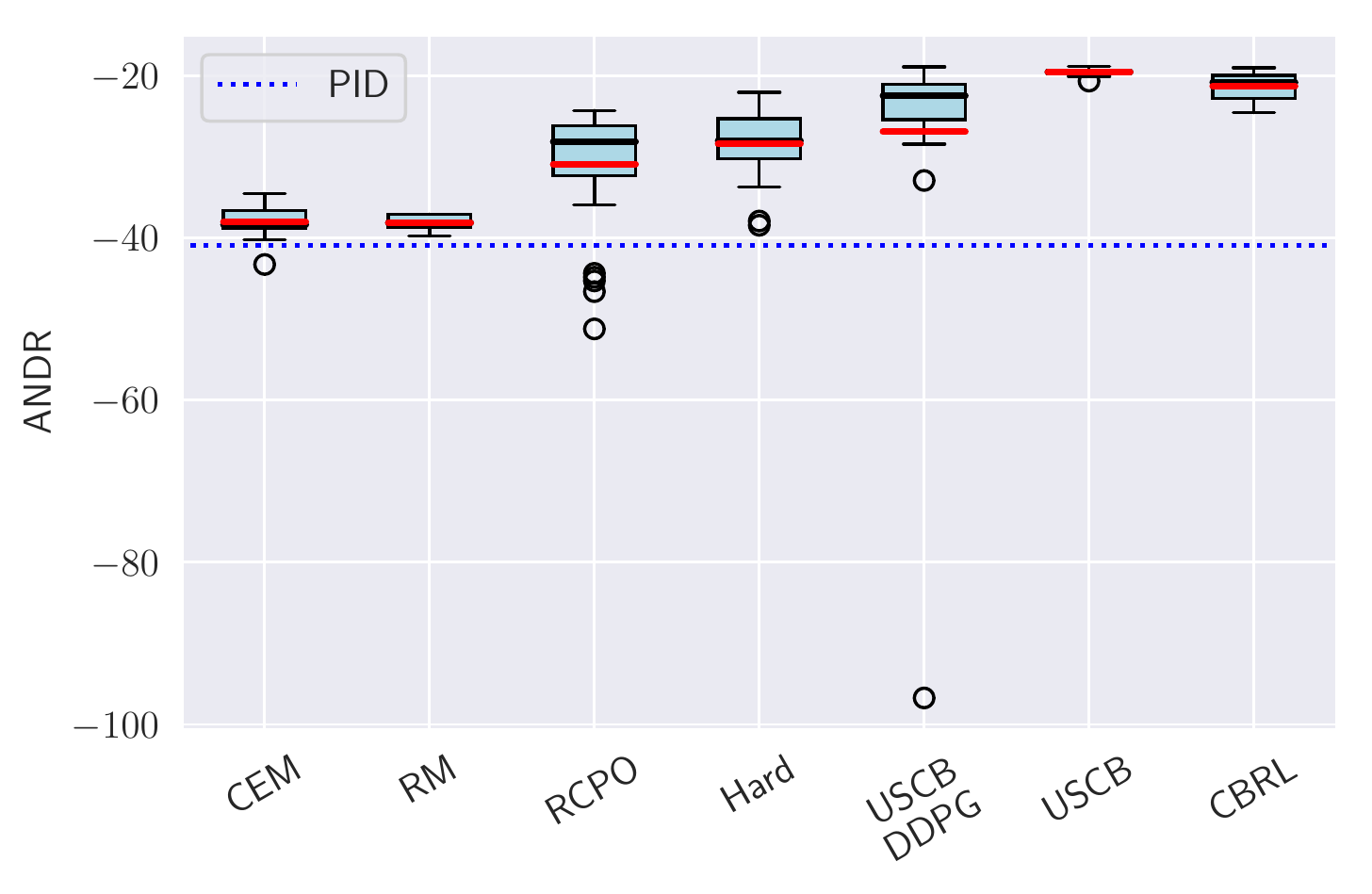}
    \end{subfigure}

    \begin{subfigure}[b]{0.3\textwidth}
        \centering
        \includegraphics[width=\textwidth]{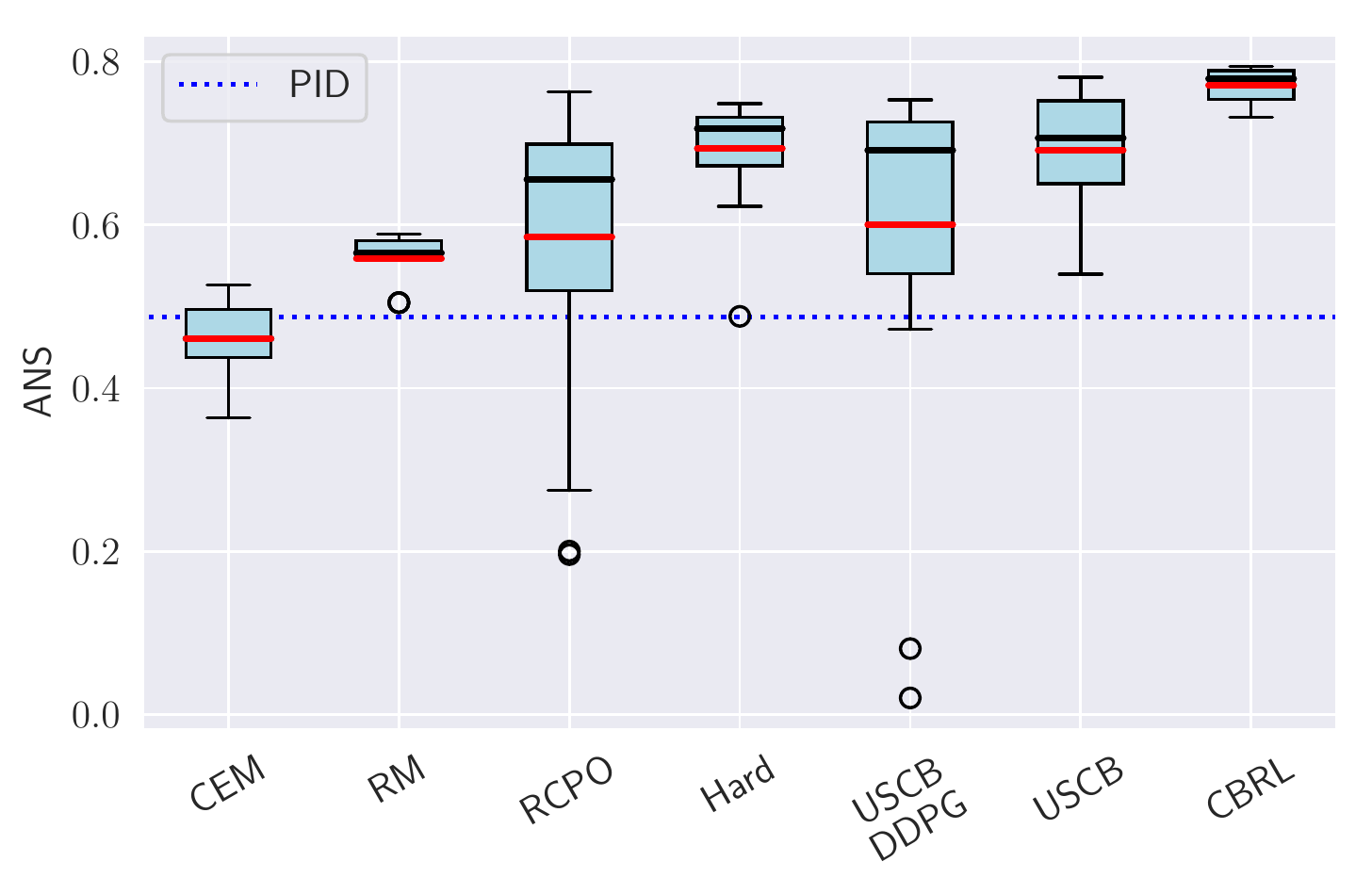}
    \end{subfigure}
    \hfill
    \begin{subfigure}[b]{0.3\textwidth}
        \centering
        \includegraphics[width=\textwidth]{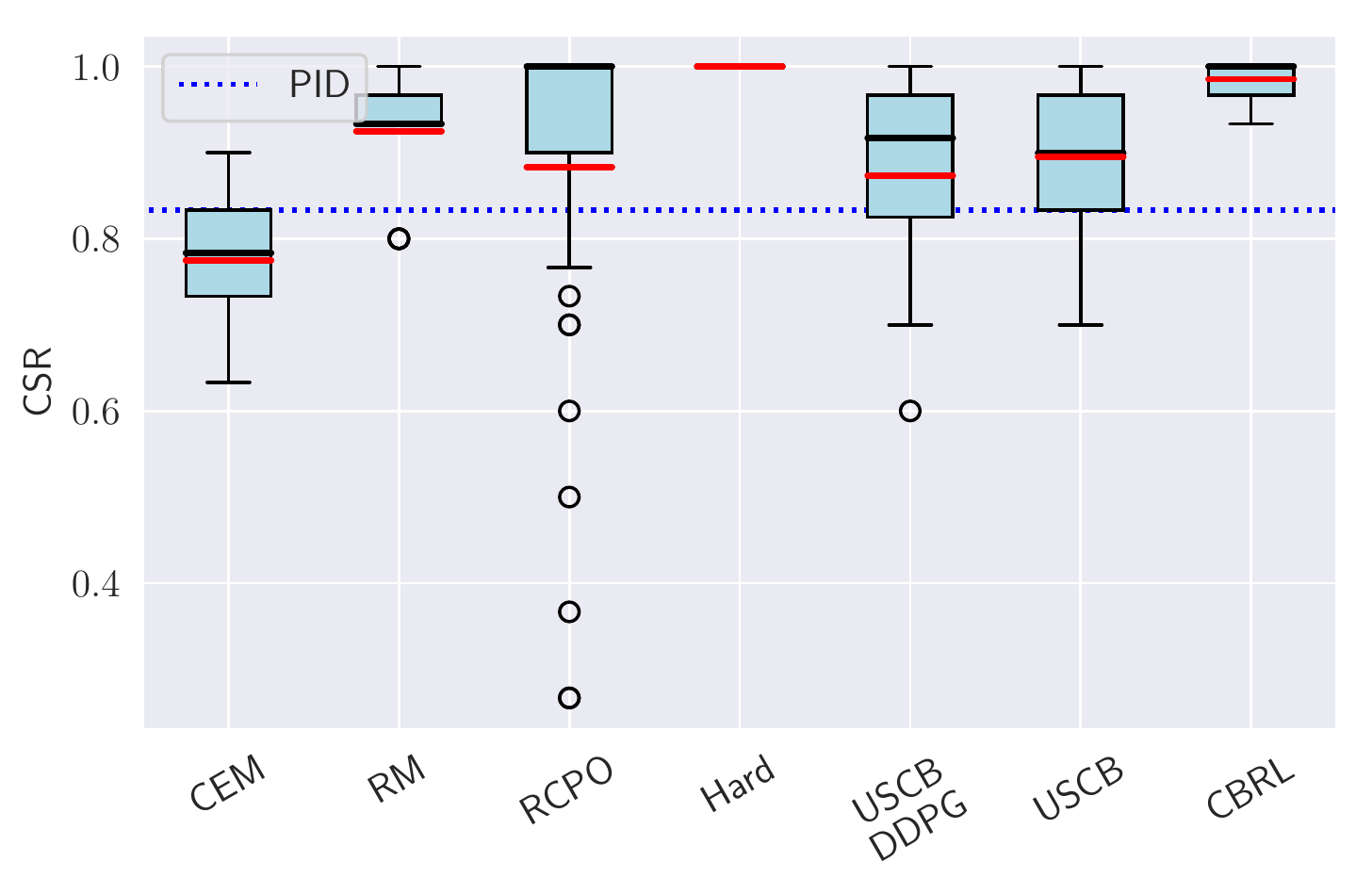}
    \end{subfigure}
    \hfill
    \begin{subfigure}[b]{0.3\textwidth}
        \centering
        \includegraphics[width=\textwidth]{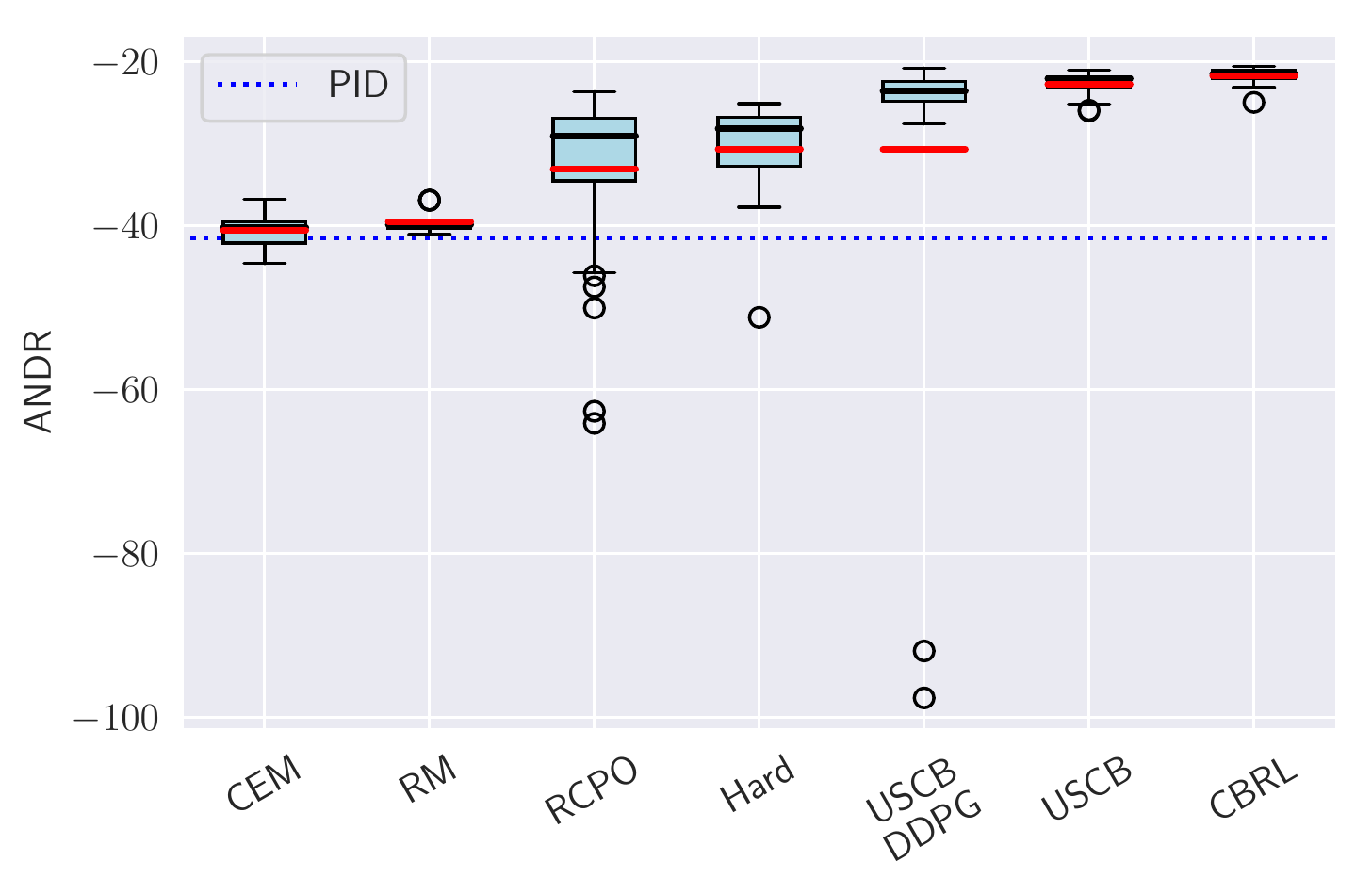}
    \end{subfigure}
    \vspace{-.3cm}
    \caption{\small \textbf{Evaluation results on the \id split.} The results of ANS (Left), CSR (Middle), and ANDR (Right) in the \sc (Top) and \mc (Bottom) settings are shown above. Each boxplot shows the average (red) and median (black) results of 20 independent repeated runs.}
    \label{fig:overall_yewu}
    \vspace{-.3cm}
\end{figure*}
}
\newcommand{\ROIDistrPlot}{
    \begin{figure}[h]
        \centering
        
        \begin{subfigure}[c]{0.46\linewidth}
             \centering
             \includegraphics[width=\textwidth]{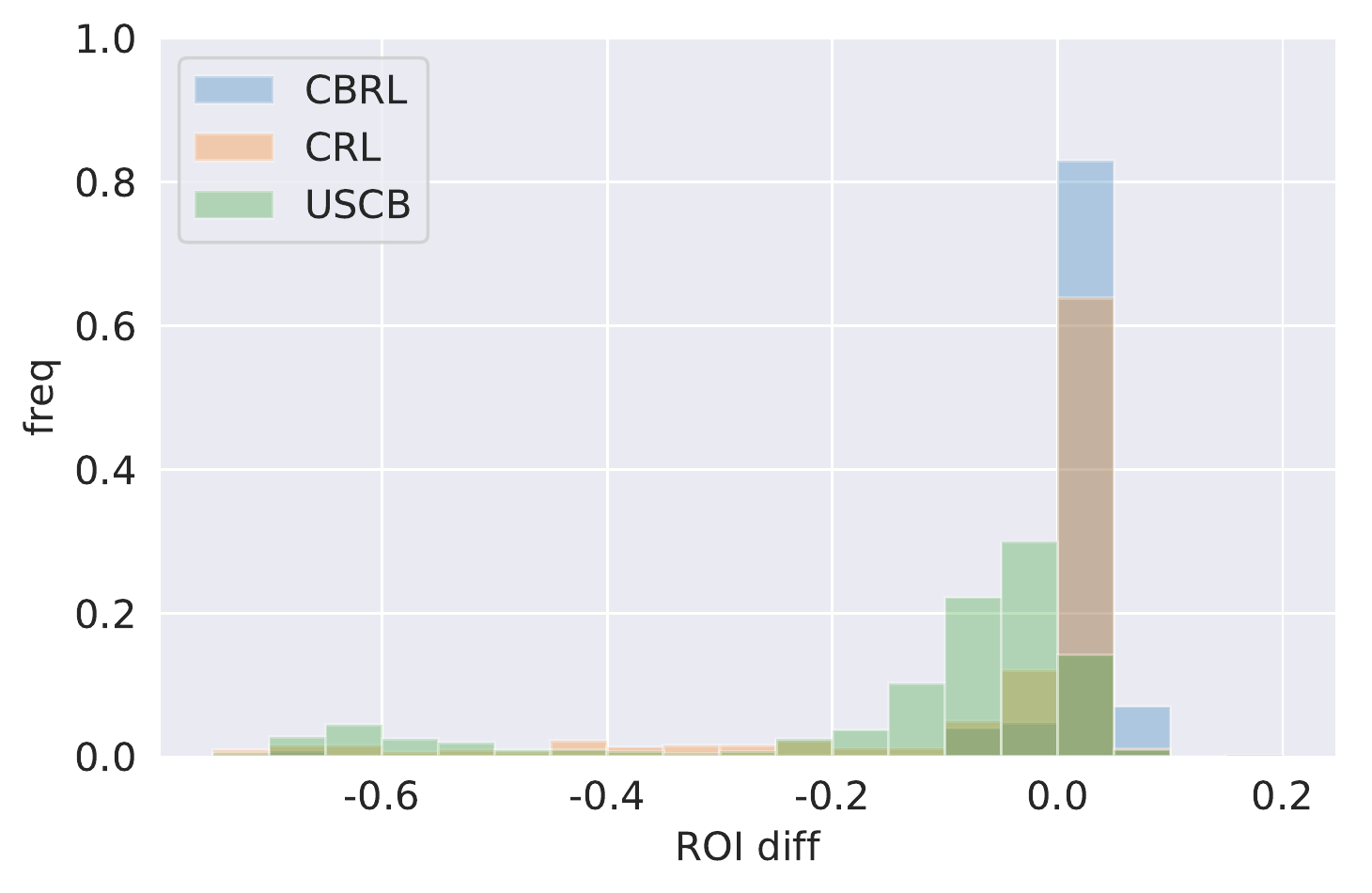}
         \end{subfigure}
         \hfill
         \begin{subfigure}[c]{0.46\linewidth}
             \centering
             \includegraphics[width=\textwidth]{sections/figs/MCB_diff_hist.pdf}
         \end{subfigure}
        \vspace{-.3cm}
        \caption{\textbf{The ROI distribution on the OOD data.} The histogram of ROI difference for models in the RCB (Left) and MCB (Right) settings of the OOD data. The x-axis is the ROI difference against the required constraint, with step 0.05. }
        \label{fig:ood_performance}
        \vspace{-.5cm}
    \end{figure}
}
\newcommand{\ROIJointPlot}{
    \begin{figure*}[t]
        \centering
        
         \begin{subfigure}[b]{0.32\linewidth}
             \centering
             \includegraphics[width=\textwidth, scale=1.5]{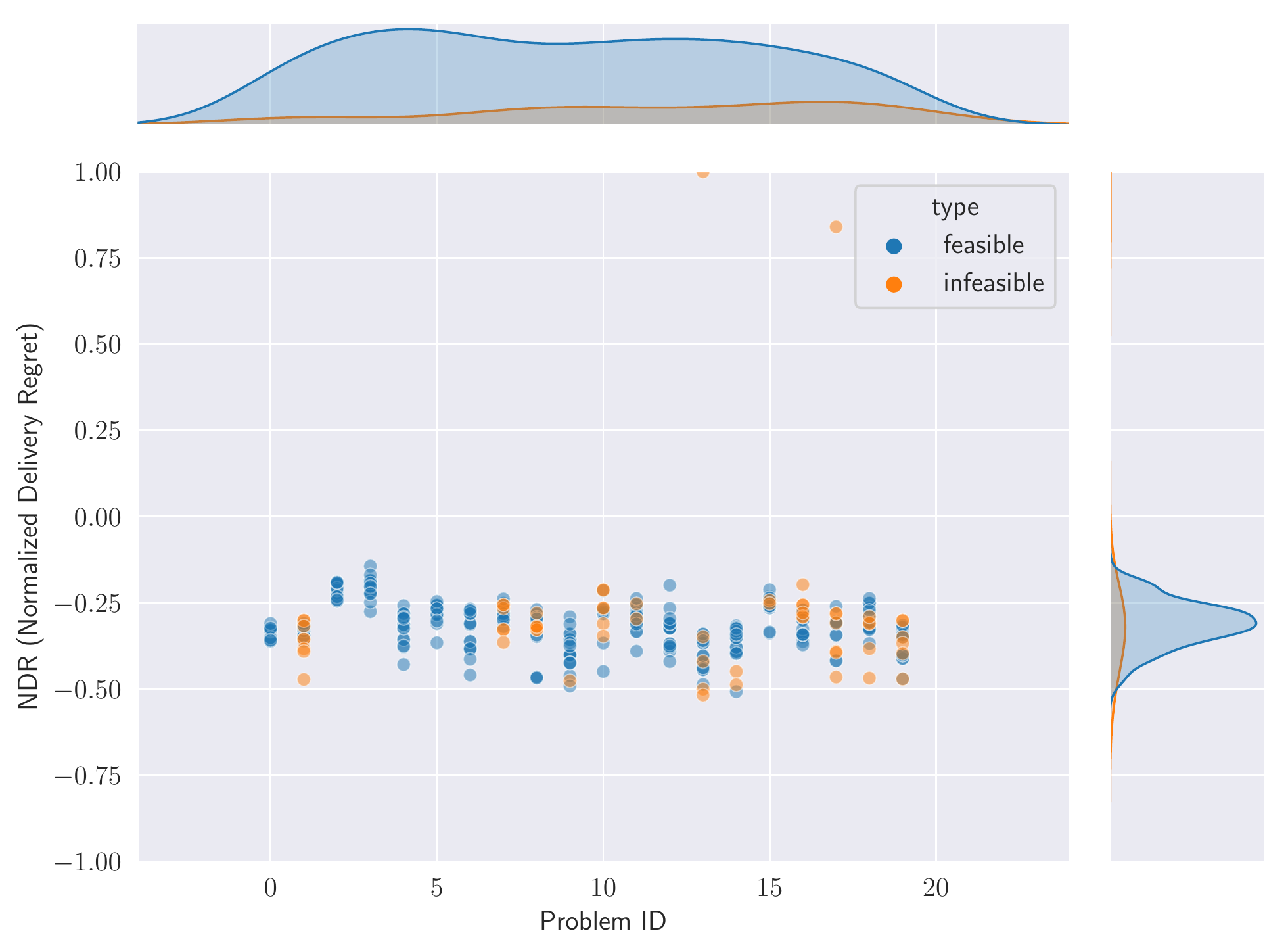}
         \end{subfigure}
        \hfill
         \begin{subfigure}[b]{0.32\linewidth}
             \centering
             \includegraphics[width=\textwidth, scale=1.5]{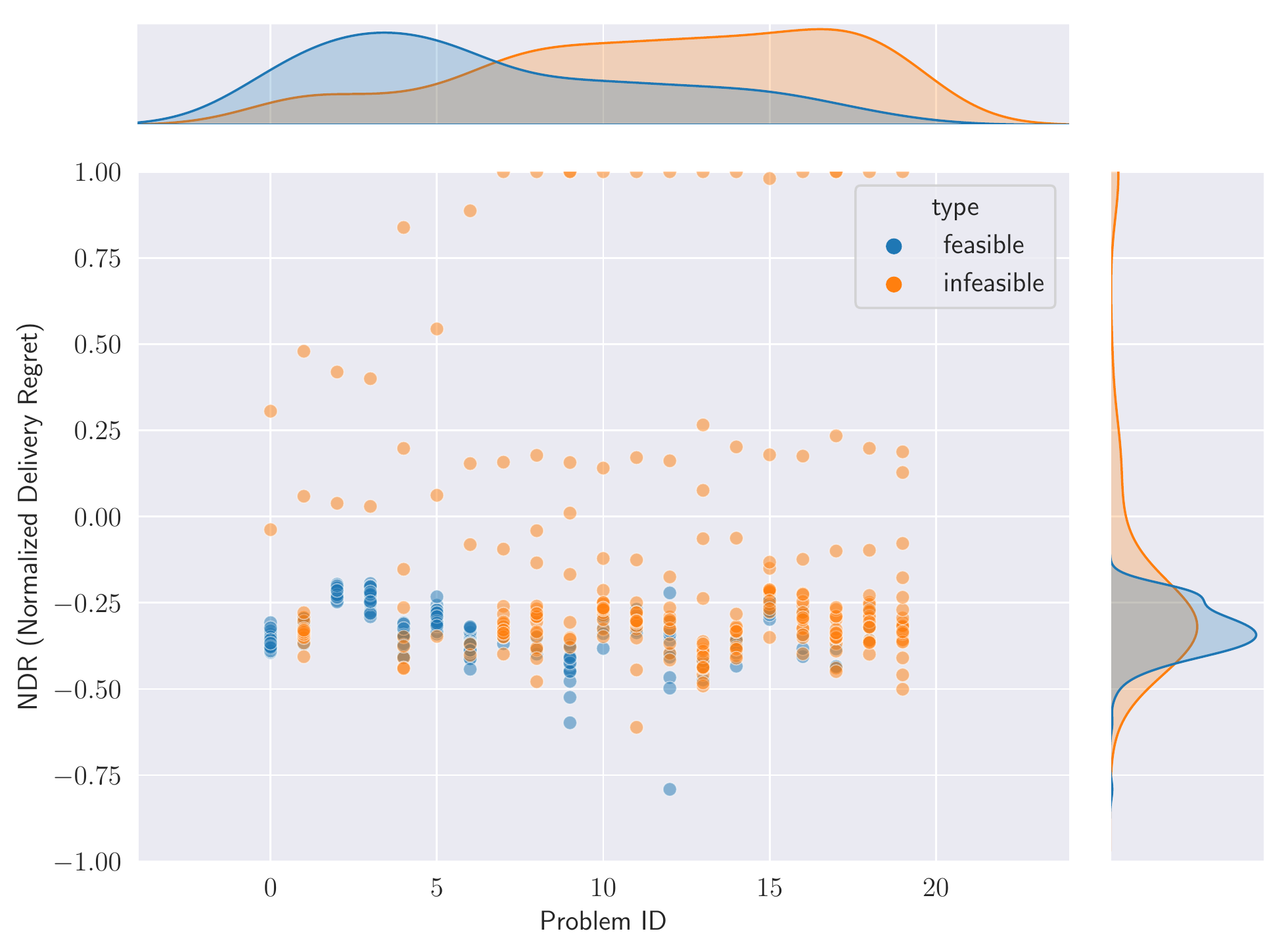}
         \end{subfigure}
         \hfill
         \begin{subfigure}[b]{0.32\linewidth}
             \centering
             \includegraphics[width=\textwidth, scale=1.5]{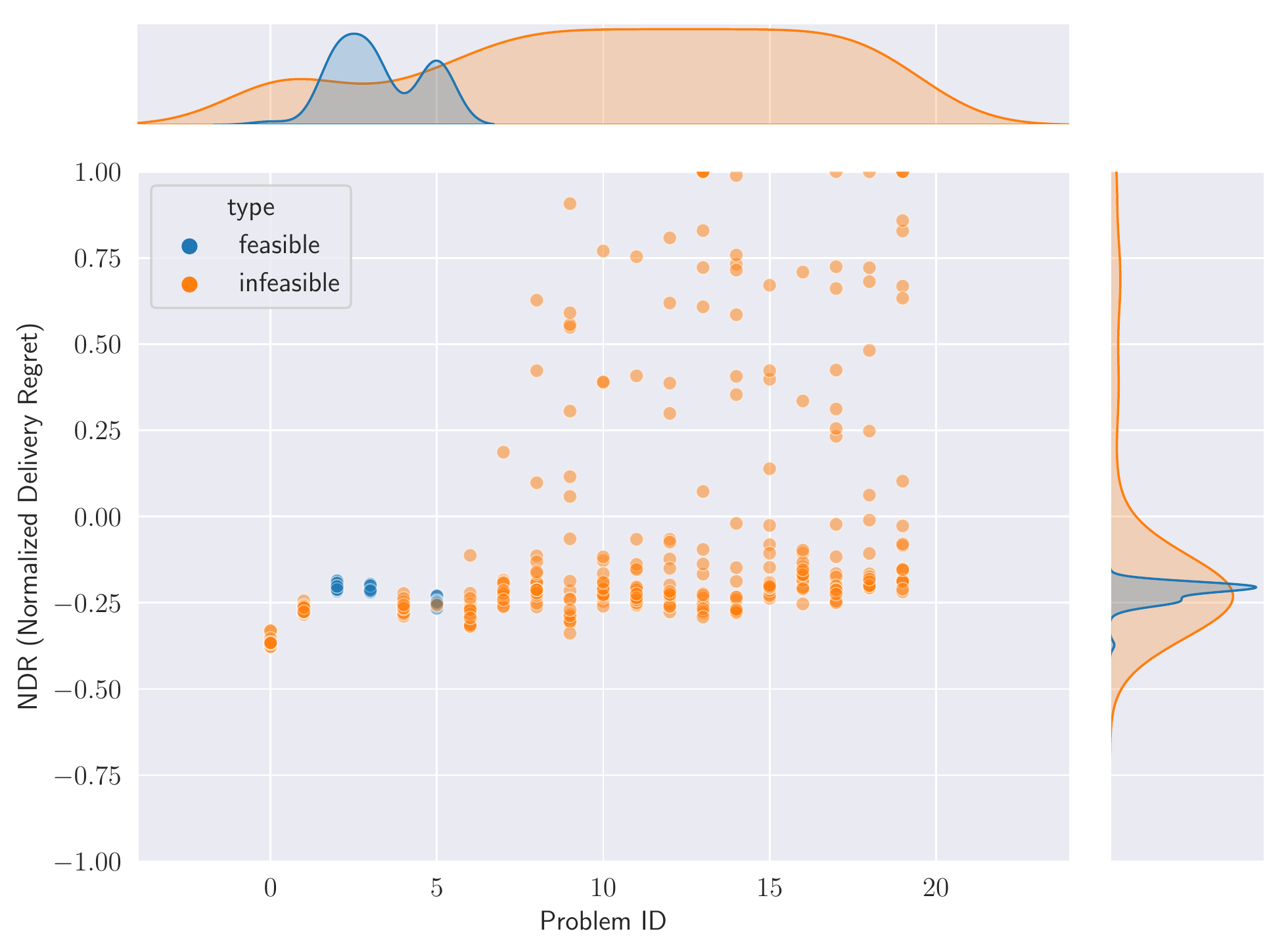}
         \end{subfigure}
         \vspace{-.3cm}
        \caption{\small \textbf{The Regret-PID joint distribution with \sc in \ood split.} Each column for CBRL, CRL, USCB (Left to Right). Each subplot contains a scatter plot for DIO and days with feasible (Blue) and infeasible (Orange) solutions, and contains the KDE plot for the marginal of regret (Right-most) and the solution distribution over problem IDs (Top-most). }
        \label{fig:ood_ours}
        \vspace{-.3cm}
    \end{figure*}
}
\newcommand{\AblPlot}{
\begin{figure*}[ht]
    \centering
    \begin{subfigure}[b]{0.3\textwidth}
        \centering
        \includegraphics[width=\textwidth]{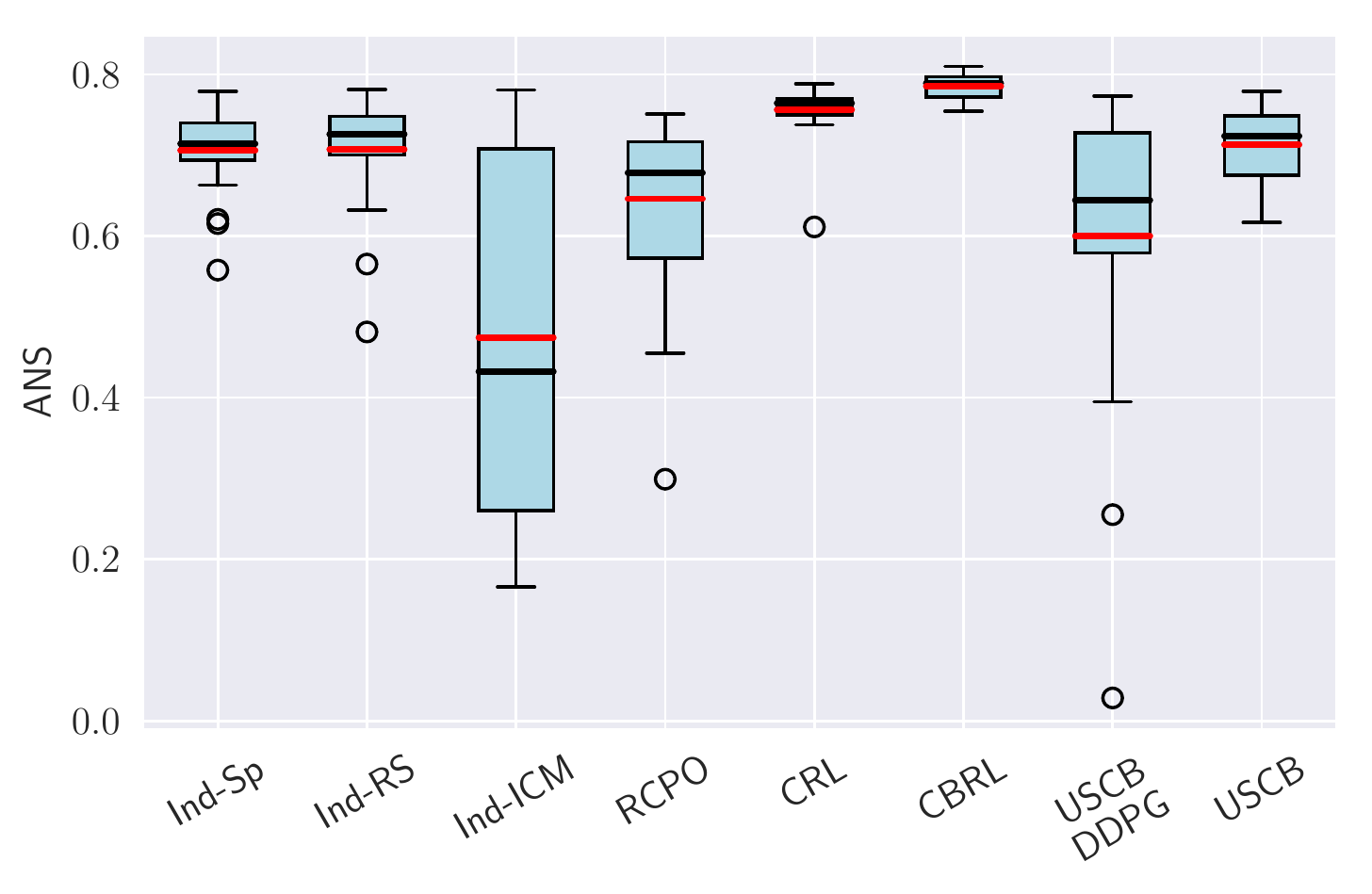}
    \end{subfigure}
    \hfill
    \begin{subfigure}[b]{0.3\textwidth}
        \centering
        \includegraphics[width=\textwidth]{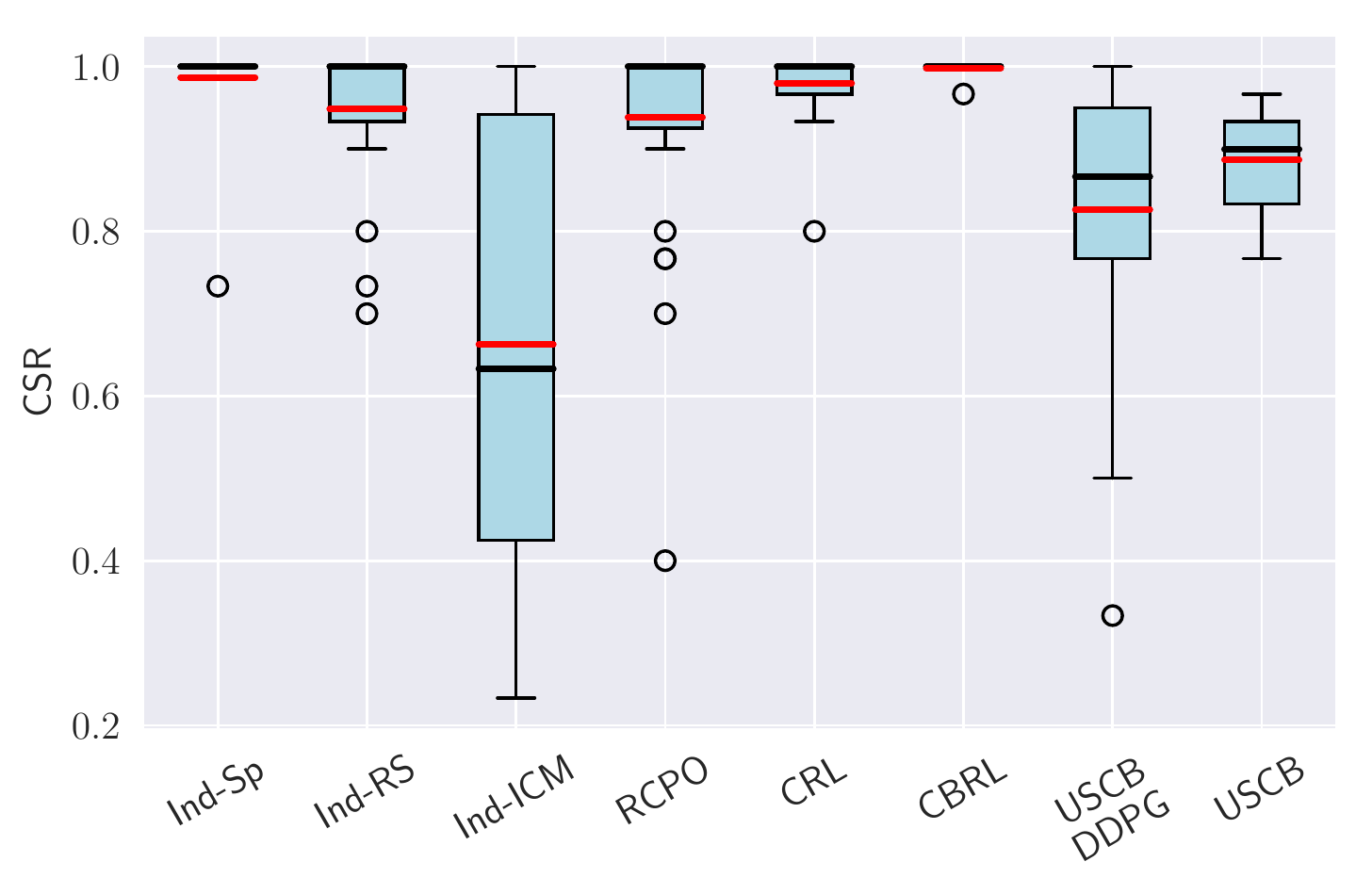}
    \end{subfigure}
    \hfill
    \begin{subfigure}[b]{0.3\textwidth}
        \centering
        \includegraphics[width=\textwidth]{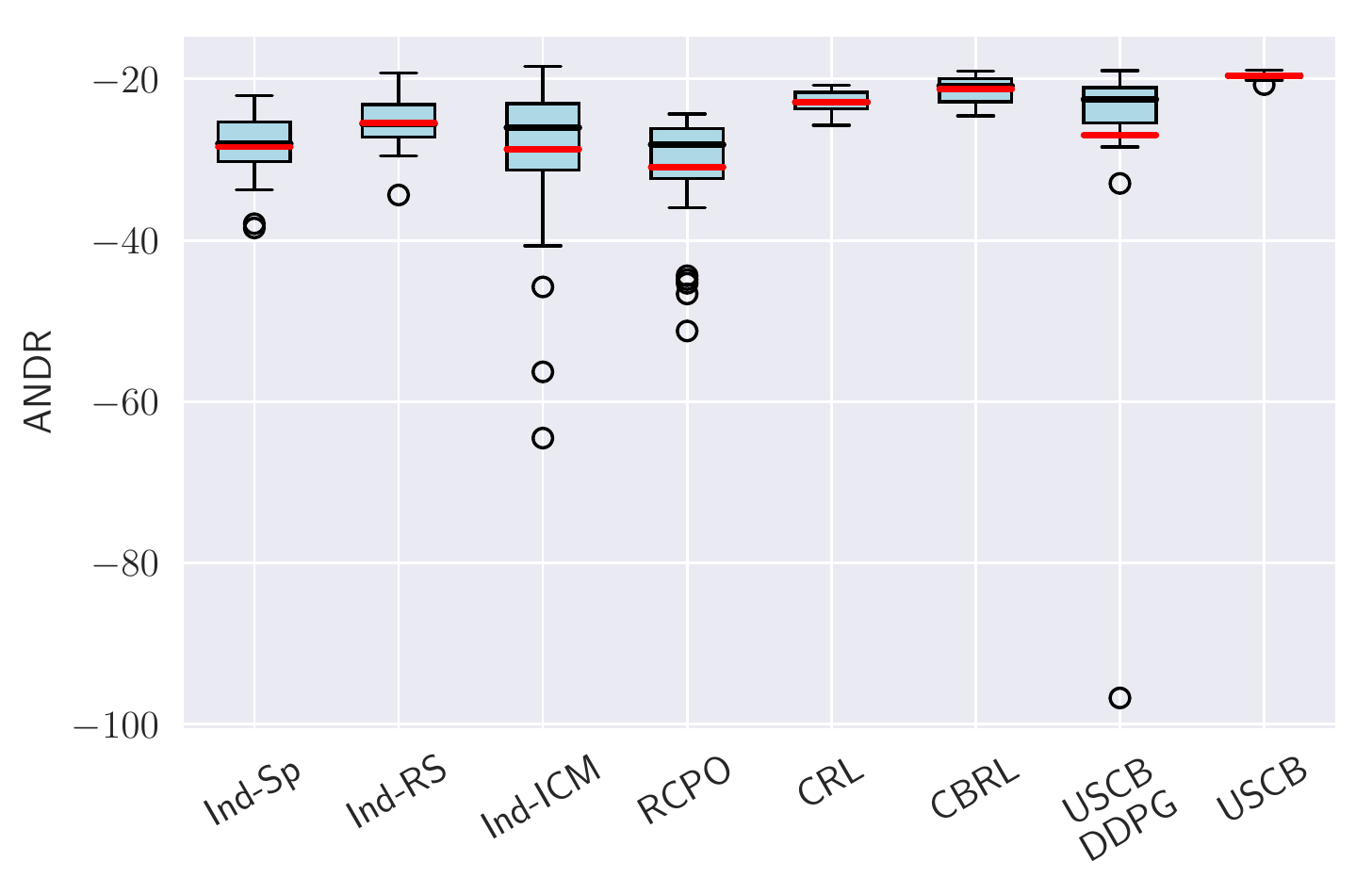}
    \end{subfigure}
    \vspace{-.3cm}
   \caption{\small \textbf{Ablations.} ANS, CSR, ANDR (Left to Right) results of ablative models evaluated in \sc.}
   \label{fig:abl_syn}
   \vspace{-.3cm}
\end{figure*}
}
\newcommand{\distrplot}{
    \begin{figure}[t]
        \centering
        \begin{subfigure}[c]{0.49\linewidth}
            \centering
            \includegraphics[width=\textwidth]{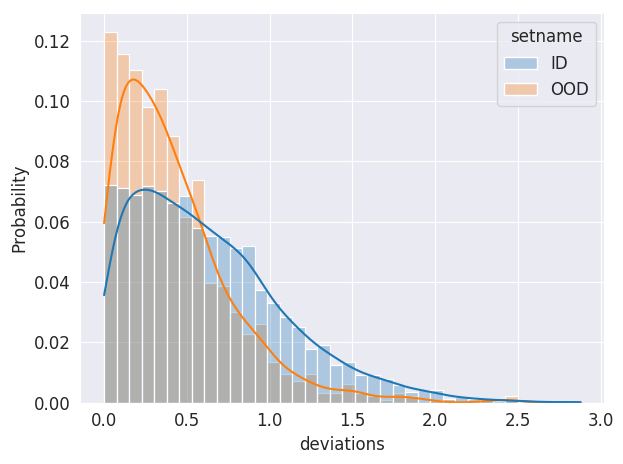}
        \end{subfigure}
        \hfill
        \begin{subfigure}[c]{0.49\linewidth}
            \centering
            \includegraphics[width=\textwidth]{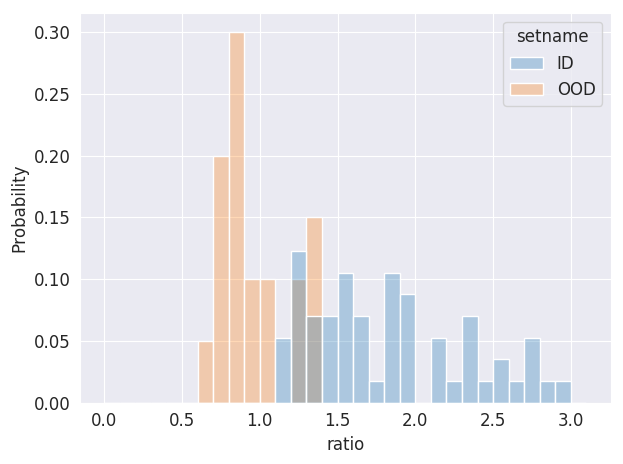}
        \end{subfigure}
        \caption{\small \textbf{Dataset.} The distribution of the slot-wise bid ratio deviations (Left) reveals the high non-stationarity of the marketplace. The distribution of the day-wise bid ratio (Right) reveals the distributional shift between \id and \ood split.}
        \label{fig:devdistr}
        \vspace{-.3cm}
    \end{figure}
    
}
\newcommand{\effplot}{
    \begin{figure}
        \centering
        \includegraphics[width=\linewidth]{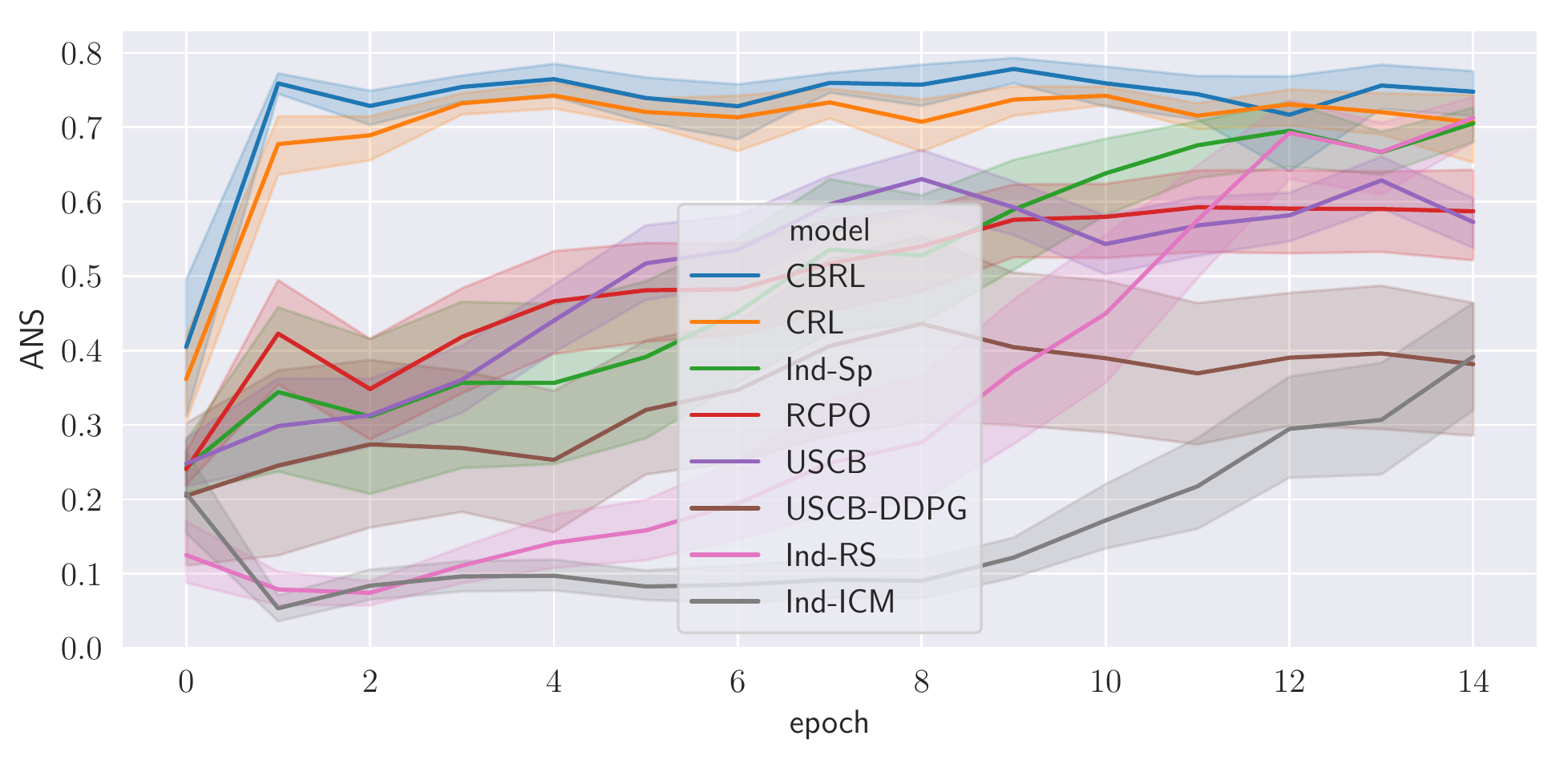}
        \vspace{-.5cm}
        \caption{\small \textbf{The learning curve on \id train set.} Confidence intervals are computed using $20$ random trials.}\label{fig:trn_eff}
        \vspace{-.3cm}
    \end{figure}
}
\distrplot
In this work, we formulate RCB in non-stationary markets as a Partially Observable Constrained MDP (POCMDP) with an \emph{indicator-augmented reward function}, and present a \emph{Curriculum-Guided Bayesian Reinforcement Learning (CBRL)} framework. Therefore, in the experiments, we aim to: (1) verify the superiority of the proposed CBRL, esp. as the first hard barrier approach to handle constraints; (2) validate the learning efficiency of the proposed curriculum learning; and (3) reveal the effects of Bayesian learning on adaptive control of the constraint-objective trade-off. We begin with a brief introduction of the evaluation setups and then detail our findings. \emph{Implementation details are left to the appendix or check out the \href{https://github.com/HaozheJasper/CBRL_KDD22}{\color{blue}{code}}.}


\vspace{-.2cm}
\subsection{Experimental Setup}
\textbf{Dataset.} We use a large-scale logged dataset from the Alibaba display advertising platform for evaluation. The dataset includes $80$-day winning bidding logs, with each day two million impressions on average. The daily logs are deliberately selected so that each day represents different market dynamics and is treated as an independent problem instance. The dataset is partitioned into two splits, containing the $60$ days and $20$ days respectively. The data distribution of the two splits is distinct from each other, with market dynamics changes caused by e-Commerce festivals, adversaries, and ad system changes (Fig.~\ref{fig:devdistr}). Consequently, the $60-$day split is termed \emph{in-distribution (\textbf{ID})}, while the $20-$day split is termed \emph{out-of-distribution (\textbf{OOD})}. We i.i.d sample from the \id  split to construct our \id train set and the \id test set with $30$ days each, which follows the conventional machine learning assumptions\footnote{Previous works~\cite{ortb, BCB,MCB} typically use week-wise logged data for training and one consecutive day for testing. We construct our benchmark different from the previous practice to obtain evaluation results that are more statistically reliable.}. In addition, the \ood split serves as an extra test set for OOD generalization, and reflects the performance of online serving after extended periods.

\noindent\textbf{Problem settings.} We construct two problem settings for the demands of platforms and performance advertisers. In the single constraint (\textbf{SC}) setting, each problem instance has a single and fixed ROI constraint $L=1$, with no budget constraint. This mimics the platform's demands to optimize social welfare in a holistic view regardless of individual budgets. In the multiple constraints (\textbf{MC}) setting, each problem instance is subject to different budget and ROI requirements, as performance advertisers expect to optimize the effects of different ad campaigns over different time periods, with different budgets and performance efficiency requirements. 

\SotaPlot
\noindent\textbf{Evaluation protocols.} In the experiment, budget constraints are satisfied surely by an early termination~\cite{BCB,MCB}, which terminates the bidding process if the budget runs out. For RCB, we introduce the following three metrics to account for overall performance, constraint satisfaction, and objective optimization\footnote{We use the metrics different from USCB~\cite{MCB} as theirs depend on an extra hyper-parameter. We include results using their metric in the appendix.}: (1) \emph{Average Normalized Score (ANS)} computes the daily score $D_T~\1_
{F}$ normalized by the oracle value $D^\ast_T$ and takes average; (2) \emph{Constraint Satisfaction Rate (CSR)} counts the ratio of feasible solutions to all problems; and (3) \emph{Average Normalized Delivery Regret (ANDR)} normalizes the delivery regret (the difference of actual delivery and best possible delivery) by the oracle delivery $D^\ast$ on condition of feasibility, and takes average. The formulas are included in the appendix. 

\vspace{-.2cm}
\subsection{Empirical Results}\label{sec:sota}
\textbf{Competing Methods.} 
We compare recent methods that can (be adapted to) deal with non-monotonic constraints, categorized into three lines of works: (1) primal-dual algorithms \textbf{RM}~(\cite{RM}); (2) slot-wise approximation methods \textbf{PID}~(\cite{PID}) and \textbf{CEM}~\cite{CEM}; (3) soft combination (RL-based) algorithms \textbf{RCPO}~(\cite{RCPO}) and \textbf{USCB} (\cite{MCB}). In addition, we include a baseline model \textbf{Hard} (same as \textbf{Ind-Sp} in the following passages), as the hard counterpart to the soft combination method \textbf{RCPO}. We provide a brief introduction for each method along with the analysis of the results below.

\noindent\textbf{Results.}
The evaluation results are shown in Fig.~\ref{fig:overall_yewu}. On both settings, \textbf{CBRL} achieves the best overall performance with stable performance (narrowest box), and performs no worse than other competing methods regarding constraint satisfaction and objective maximization. Specifically, in the \sc setting, \textbf{CBRL} reaches 78.9\% of the oracle (median ANS 0.789), satisfies the constraints almost surely (median CSR 1), with feasible solutions falling behind oracle by 20.8\% (median ANDR 0.792)\footnote{Note that we use slot-wise oracle policy which reaches higher performance upper bound than in previous studies \cite{MCB}. Check the appendix for details.}. Similar results are obtained in \mc, with median ANS $0.789$, median CSR 1.0, and median ANDR -21.5\%.
The results of competing methods are analyzed as 
follows. 
\begin{itemize}[leftmargin=*]
    \item \textbf{RCPO}~(\citeyear{RCPO}) is a general CMDP policy optimization algorithm based on Lagrangian relaxation, which introduces extra \emph{learning parameters} to control the constraint-objective trade-off. We adapt RCPO to RCB by the proposed POCMDP formulation. While its maximal performance reaches \cbrl, its average performance is hampered by high variance. The instability is due to (1) the sensitive initializations of the Lagrangian multipliers for stochastic optimization, especially in the non-convex condition; and (2) the soft combination of constraint violations and delivery with improper weights that leads to ambiguity in rewards. Particularly, we notice that the ambiguity makes instability combinatorially more likely in \mc than in \sc (wider box in \mc than in \sc).
    \item \textbf{USCB}~(\citeyear{MCB}) is the prior art that formulates an RL framework and uses extra \emph{hyper-parameters} to non-linearly control the constraint-objective trade-off, which can be treated as a variant of Lagrangian relaxation. For a fair comparison, our method CBRL and USCB use the same input features and network structures. We provide two specific implementations, \uscb and \textbf{USCB-DDPG}. \uscb aligns with CBRL in entropy regularization~\cite{sac} and independent action space, while \textbf{USCB-DDPG} respects the plain Monte Carlo estimation based actor-critic approach and the temporally correlated action space in \cite{MCB} (check the appendix).
    
    It follows that \uscb tends to be more stable than \textbf{RCPO} (narrower box), as the instability caused by non-convex optimization is alleviated.  However \uscb still suffers from the reward ambiguity. In particular, we remark that while the best USCB model (top-rating in ANS) exhibits the best ANDR performance (at the cost of constraint satisfaction), its trade-off parameter design indeed shows a significant see-saw effect, and requires laborious tuning. By contrast, \cbrl adopts a parameter-free solution, which is user-friendly and turns out best-performing in ANS.
    \item \textbf{CEM}~\cite{CEM} Cross-Entropy Method is a gradient-free stochastic optimization method. Widely used in the industry, CEM attempts to optimize a greedy sub-problem in each time slot and bears the exploration-exploitation trade-off. Since winning is sparse in the data, more exploration is required to obtain a more accurate estimate, which squeezes the space for exploitation. Consequently, the best CEM model achieves decent constraint satisfaction (around 0.8 in CSR) but lower objective optimization, due to the averaging effect of the dominant exploration traffic.

    \item \textbf{PID}~(\citeyear{PID}) adopts a PID control solution to bidding with CPC constraint and budget constraint. Based on the optimal bidding function~\eqref{optb}, we adapt \textbf{PID} to control the bid ratio that drives the ROI constraint toward the target in each time slot. We note that PID itself does not handle changing systems well, and the online adjustment of PID parameters to suit the changed systems is non-trivial and beyond the scope of this paper. We find empirically that PID cannot balance constraint-objective trade-off well in highly non-stationary markets presented in our dataset, and hence the best PID model with the best ANS score shows the only moderate status of both constraints satisfaction (CSR) and objective optimization (ANDR). 
    \item \textbf{RM}~(\citeyear{RM}) propose to deal with RCB under a static functional optimization framework, which solves the optimal bid ratio over the train set and applies to the test set. The solved bid ratio achieves the best performance on the training problems on average, but does not adapt to each of them. As a result, in the \id test set, the RM model performs far from optimal, although it respects the constraints well (CSR close to 1).
    
\end{itemize}

\effplot
\AblPlot
\ROIJointPlot
\vspace{-.2cm}
\subsection{Effects of Curriculum Learning}\label{sec:expcl}
Reward sparsity is notorious for inefficient policy learning due to blinded exploration. We propose a curriculum-guided policy search process and use a three-stage curriculum sequence in the experiments, including one cold-start curriculum for three epochs, and one warm-up curriculum for three epochs, followed by the original problem. To verify its effectiveness in promoting efficient learning, we compare with the following baseline models: (1) \textbf{Ind-Sp} uses the sparse reward function Eq.~\eqref{rprime}; (2) \textbf{Ind-RS} uses the reward shaping~\cite{rs, rscl} technique, which introduces extra reward signals in each slot based on human expertise; (3) \textbf{Ind-ICM} applies the Intrinsic Curiosity Module (ICM)~\cite{ICM}. Intuitively, ICM uses reconstruction errors as extra reward signals. 

Fig.~\ref{fig:abl_syn} shows that, on the \id test set, the proposed curriculum learning baseline \textbf{CRL} outperforms reward shaping and ICM, both of which even perform worse than the sparse reward baseline. Learning curves of the baseline models are shown in Fig.~\ref{fig:trn_eff}, indicating \textbf{CRL} achieves high performance within three epochs (the first curriculum), which already beats the prior art \uscb.  It can be concluded that curriculum learning (\textbf{CRL}) improves training efficiency from sparse rewards (\textbf{Ind-Sp}), while the extra signals provided by reward shaping and ICM appear noisy to the policy at the beginning (downward curve) and afterward cause instability during learning (high error band). 

\vspace{-.2cm}
\subsection{Effects of Bayesian Learning}\label{sec:expbl}
In the proposed hard barrier solution, Bayesian learning takes the responsibility to adaptively control the constraint-objective trade-off according to market dynamics. Here we examine two scenarios, in-distribution (\textbf{ID}) and out-of-distribution (\textbf{OOD}) data regimes. 

Fig.~\ref{fig:abl_syn} shows, \cbrl improves \textbf{CRL} in \id split (median ANS from $0.764$ to $0.789$). More significant improvement is witnessed in the challenging \ood split (median ANS from 0.24 to 0.54, see the appendix). In Fig.~\ref{fig:ood_ours}, the feasible solution distributions (top KDE plots) show \cbrl achieves better constraint satisfaction than \textbf{CRL} (median CSR 0.775 v.s. 0.35), while \uscb fails dramatically (median CSR 0.15). Among the feasible solutions, the regret distribution (right-most KDE plots) show both \cbrl and \textbf{CRL} deteriorates by 10\% in regret from the \id split (median ANDR -30.18\% v.s. -32.93\%). As \cbrl is not specifically designed for \ood, such performance is sub-optimal yet reasonable. Moreover, the regret does not increase much while respecting constraints, indicating \cbrl achieves good constraint satisfaction not through a (too) conservative policy, but through market adaptability. The scatter plots show the specific failure patterns in \ood split. We remark that, compared with \textbf{CRL}, \cbrl rarely presents orange points high above, which earns excessive delivery (regret of feasible solutions are below 0) at the cost of severely violated constraints. 

The good performance in \ood scenario is credited to Bayesian learning. Logically, although the joint data distribution is different between \id and \ood split, the distribution of market variations can overlap. It follows that the variational distribution $q(z)$ may generalize (partially) to \ood. Since the agent expresses its uncertainty about the market by $q(z)$, and eliminates the uncertainty through the iterative process of posterior sampling, correctly inferred posterior $q(z)$ empowers the agent to achieve adaptive control of the constraint-objective trade-off in the unseen environment.

\vspace{-.2cm}
\section{Related Work}\label{sec:related}
\vspace{-.1cm}


\noindent\textbf{Reward Sparsity.}
Learning with sparse reward is challenging since RL relies on the feedback from the environment to narrow down the search space for the optimal policy. Reward shaping~\cite{rs,rscl} relies on expertise and suffers from biased learning. Intrinsic motivations have been proposed to help exploration with extra forces, e.g., surprise~\cite{ICM}. Curriculum Learning has been explored on supervised learning~\cite{cl2}, and is extended to RL in \cite{cl, rscl} to deal with sparse reward. In this work, we exploit the problem structure of constrained bidding for efficient policy learning. 
\noindent\textbf{Bayesian RL.}
Policy learning in unknown environments requires to balance the exploration-exploitation trade-off. A Bayes-optimal policy does this optimally by conditioning on not only the observations but also agent's uncertainty about the current MDP. Under the framework of \emph{Bayes Adaptive Markov Decision Processes (BAMDP)}, a policy is Bayes-optimal, by augmenting the state space with a belief distribution over possible MDPs~\cite{BAMDP}. The idea is associated with belief MDP~\cite{pomdp2} that represent POMDPs as a belief over MDPs. To achieve tractability, planning on BAMDP is usually achieved by posterior sampling~\cite{strens, osband} in an MDP periodically sampled from the hypothesis over MDPs. Following this, works in solving POMDPs~\cite{dvrl} or meta-RL problems~\cite{varibad,taskinf, caster}, learn approximately Bayes-optimal policies while maintaining a posterior distribution over MDPs, usually via deep variational inference~\cite{dvl,blei2017variational}. In this work, we adopt similar ideas to achieve adaptive control in partially observable non-stationary markets.

\noindent\textbf{Constrained RL.}
Various methods~\cite{RCPO,achiam2017constrained,chow2017risk,chow2019lyapunov} have been proposed to solve CMDPs. Lagrangian relaxation is commonly adopted in \cite{RCPO,chow2017risk}, which introduces Lagrangian multipliers to control the constraint-objective trade-off, and is shown to have stability issues~\cite{chow2019lyapunov}. RCPO~\cite{RCPO} relates with our method closely as we both accomodate constraints into the reward function. However, the proposed indicator augmention method is parameter-free and exploits the problem structure in contrast to RCPO that degenerates in dynamics-varying environments.

\vspace{-.2cm}
\section{Conclusion}
\vspace{-.1cm}
In this work, we propose the first hard barrier solution to RCB. Based on a Partially Observable Constrained MDP formulation, the indicator-augmented reward function in conjunction with the Curriculum-Guided Bayesian Reinforcement Learning framework achieves adaptive control in partially observable non-stationary markets, without laborious tuning for hyper-parameters. Extensive experiments on a large-scale industrial dataset with two problem settings verify the superior generalization and stability of our method in both in-distribution and out-of-distribution data regimes.

\vspace{-.2cm}

\bibliographystyle{ACM-Reference-Format}
\bibliography{main}

\clearpage
\appendix
\newcommand{\GOverallPlot}{
\begin{figure*}[ht]
    \centering
    \begin{subfigure}[b]{0.32\textwidth}
        \includegraphics[width=\textwidth]{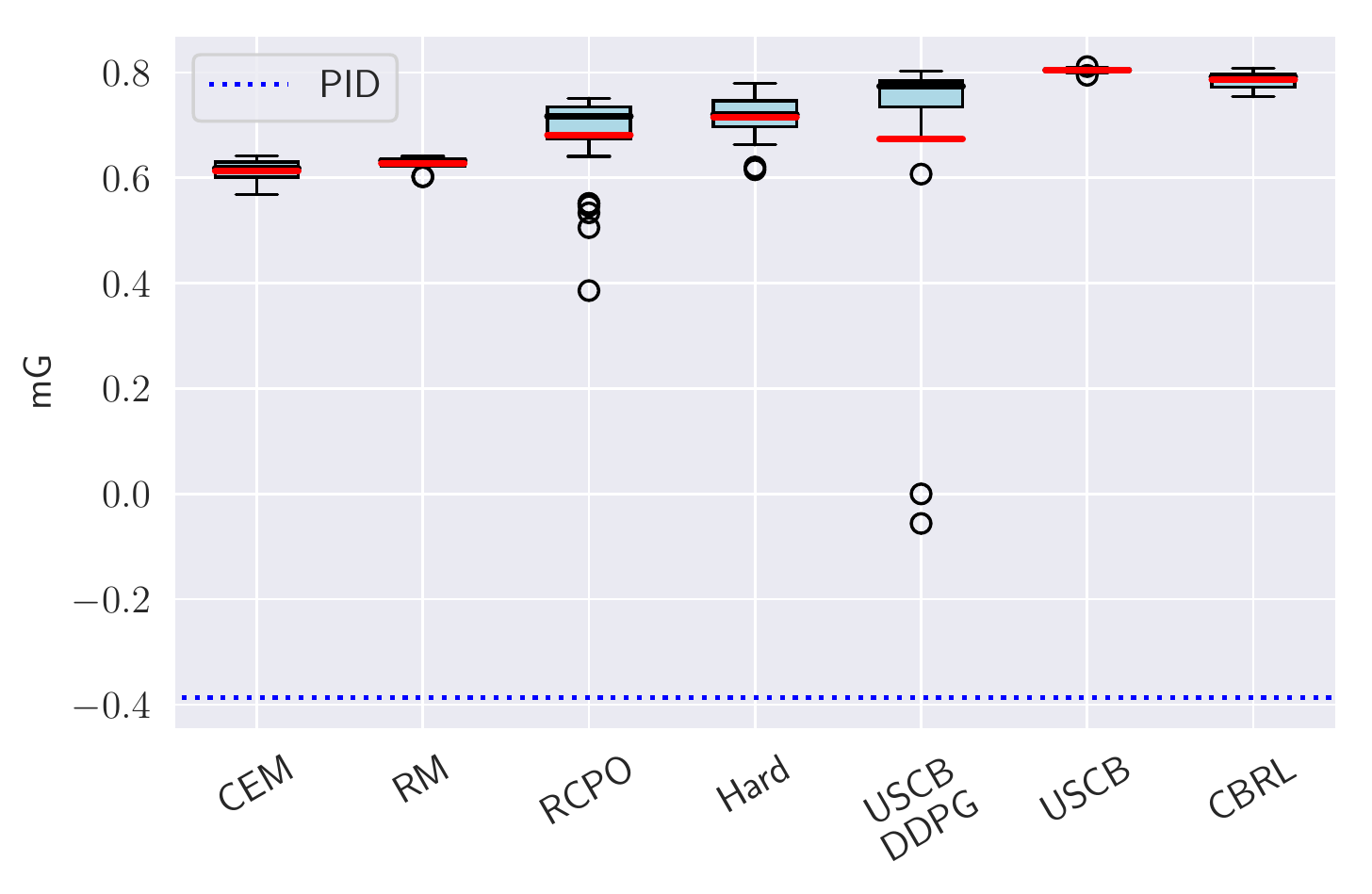}
    \end{subfigure}
    \begin{subfigure}[b]{0.32\textwidth}
        \includegraphics[width=\textwidth]{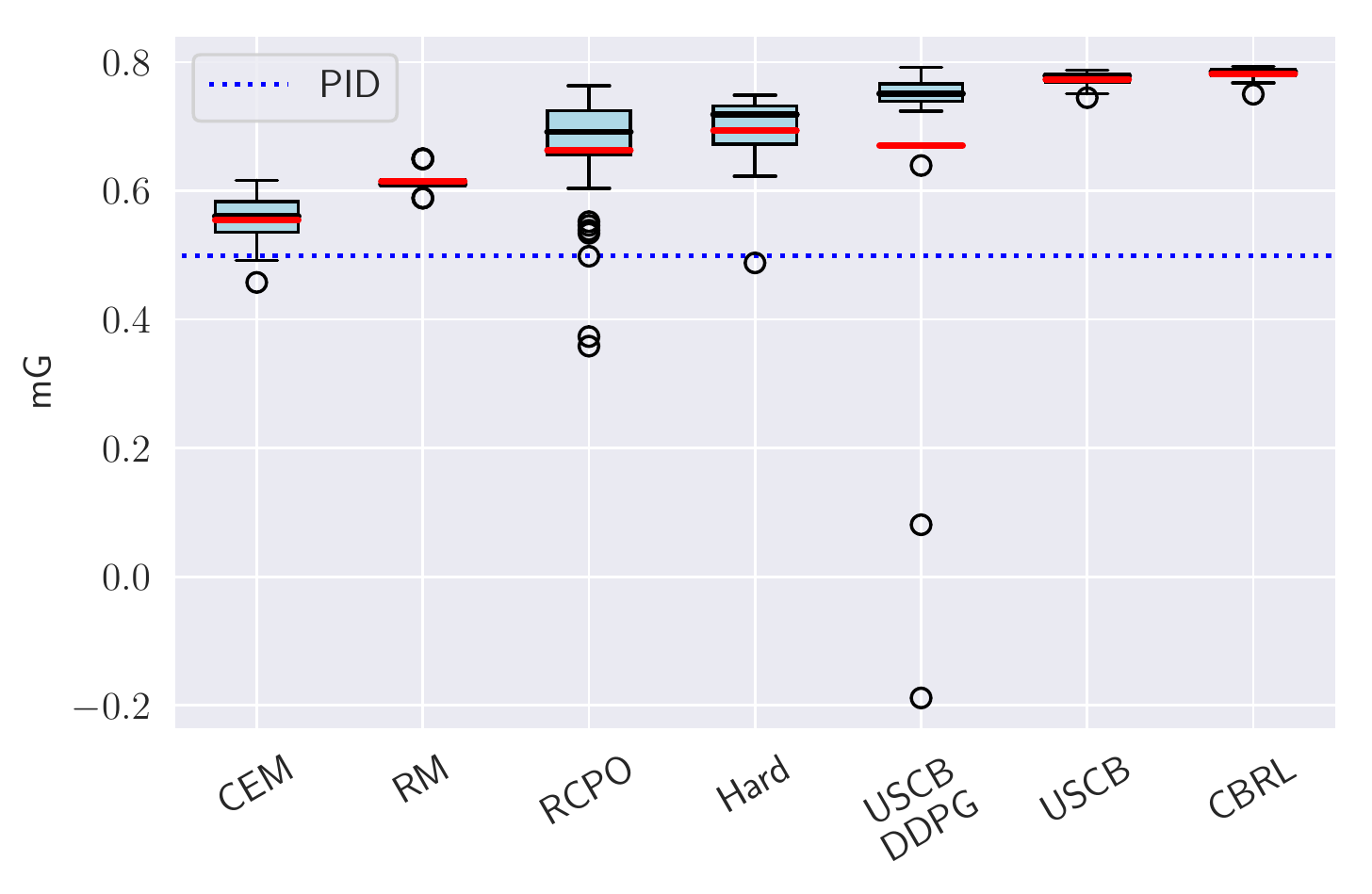}
    \end{subfigure}
    \begin{subfigure}[b]{0.32\textwidth}
        \includegraphics[width=\textwidth]{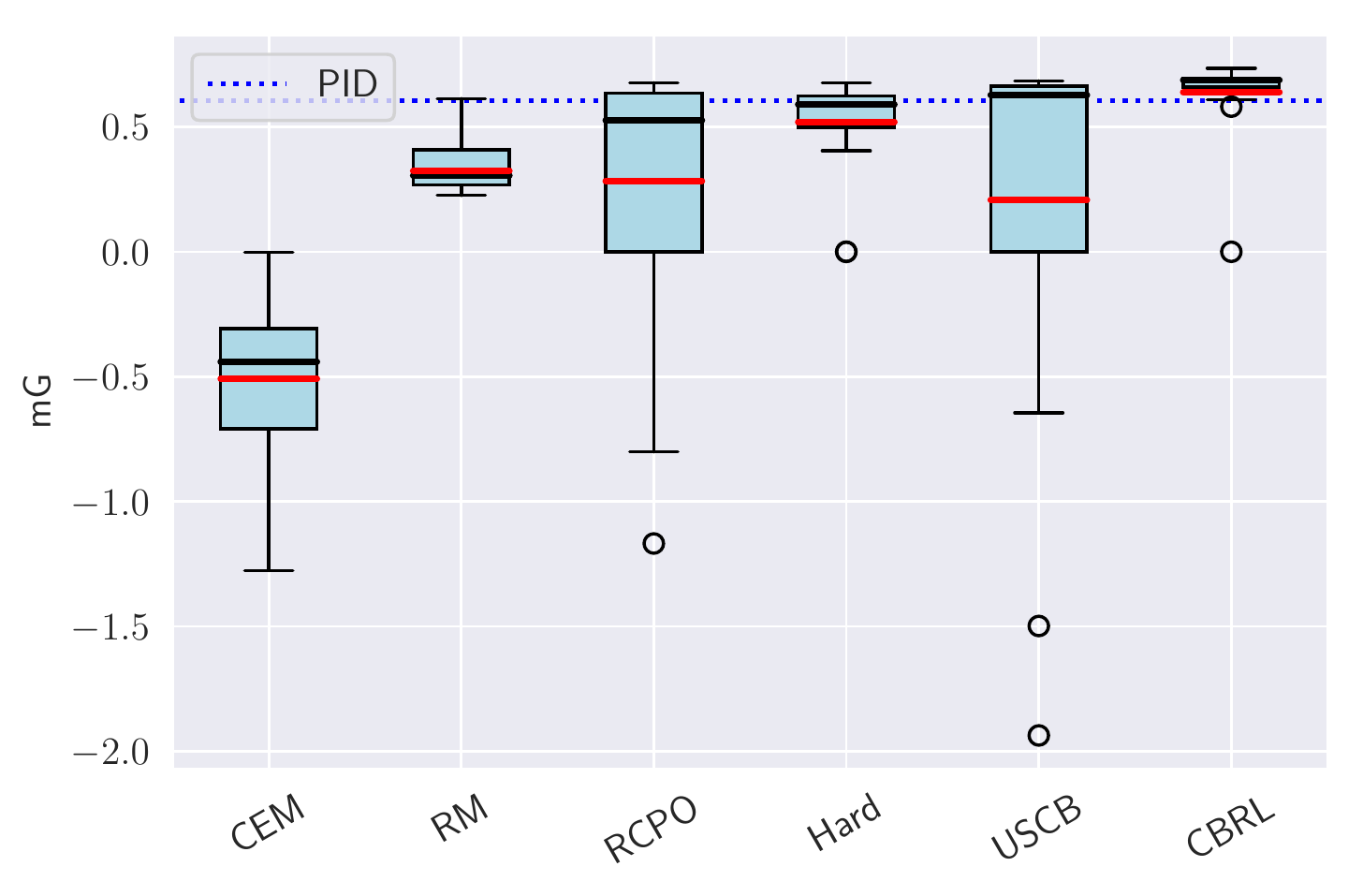}
    \end{subfigure}
    \caption{\textbf{The G-metric performance of \sc (Left) setting and \mc (Middle) setting on \id split, and \sc setting on \ood split (Right). } }\label{fig:overall_gscore}
\end{figure*}
}

\newcommand{\OODPlot}{
\begin{figure*}[ht]
    \centering
    \begin{subfigure}[b]{0.3\textwidth}
        \centering
        \includegraphics[width=\textwidth]{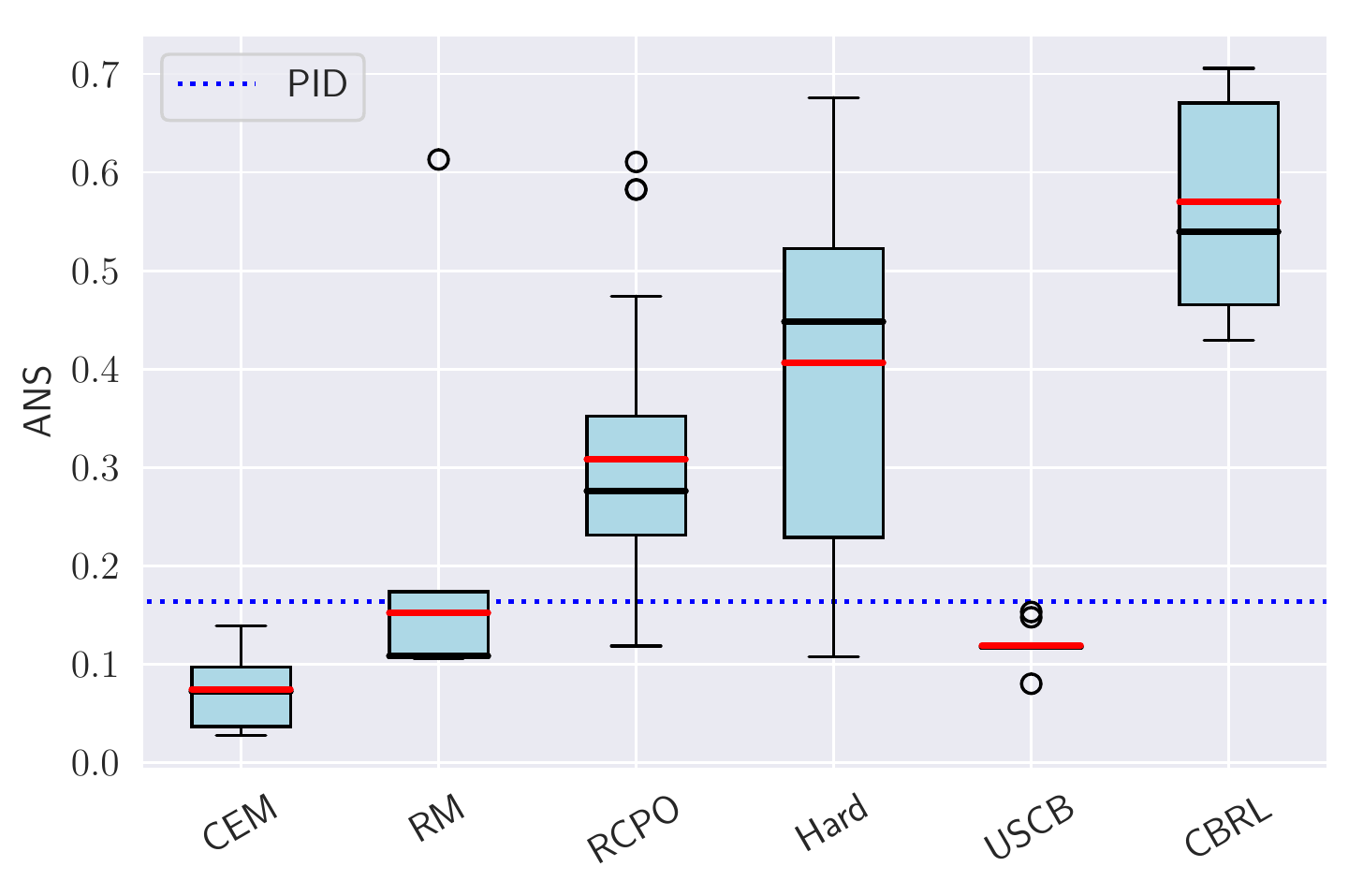}
    \end{subfigure}
    \hfill
    \begin{subfigure}[b]{0.3\textwidth}
        \centering
        \includegraphics[width=\textwidth]{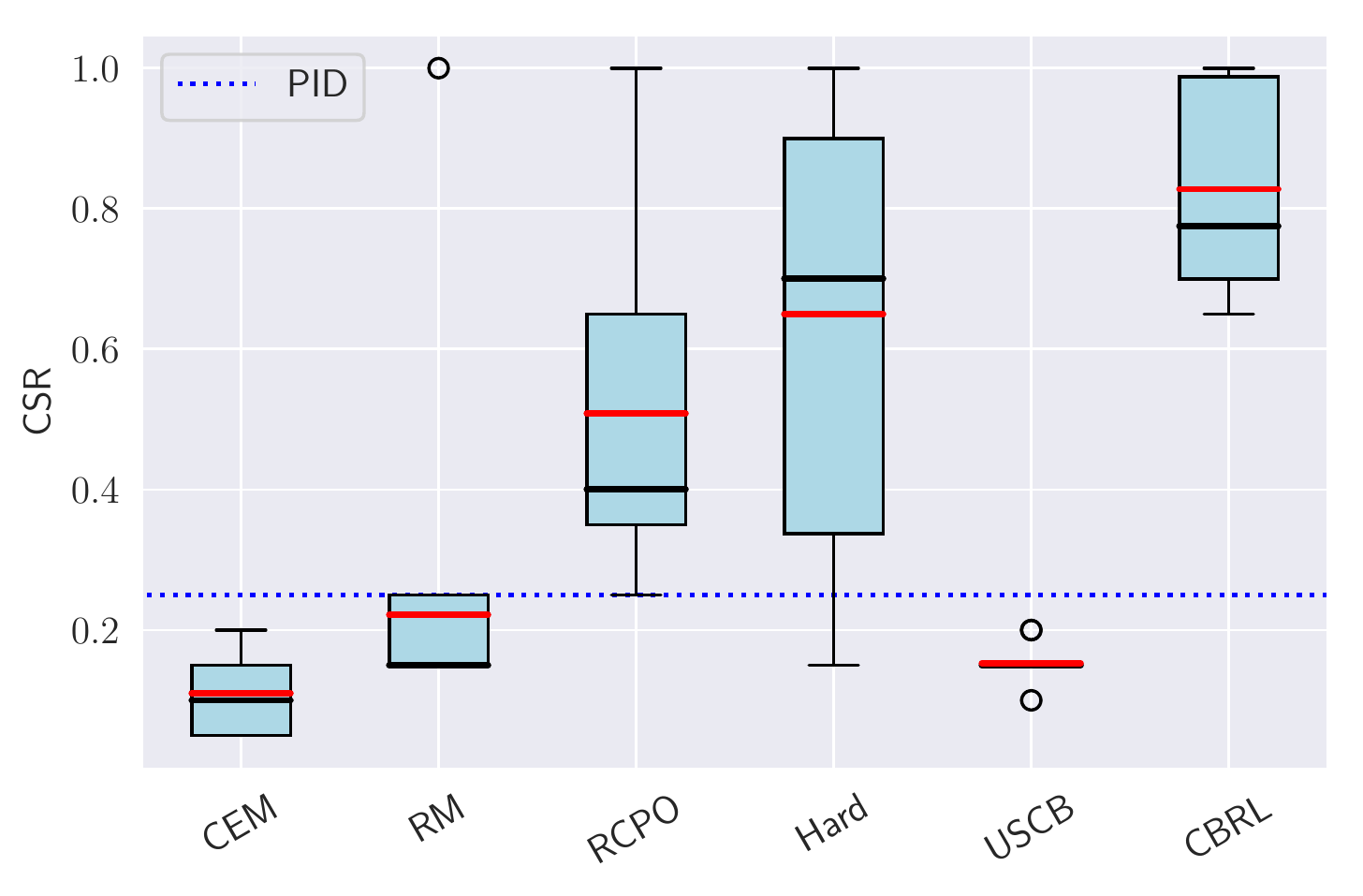}
    \end{subfigure}
    \hfill
    \begin{subfigure}[b]{0.3\textwidth}
        \centering
        \includegraphics[width=\textwidth]{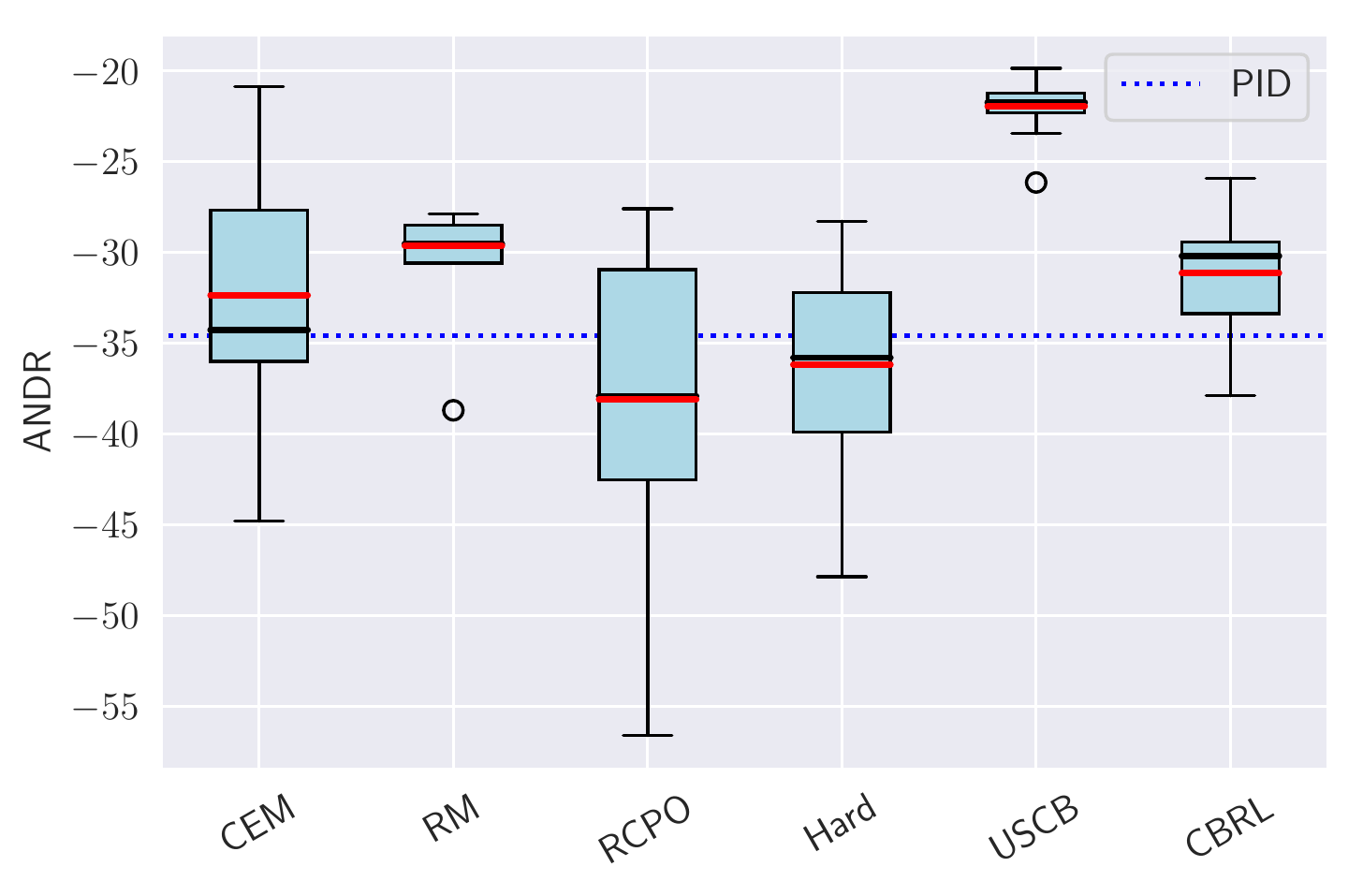}
    \end{subfigure}
    \vspace{-.3cm}
   \caption{\small \textbf{Results of competing methods on \ood split.}}\label{fig:oodplot}
   \vspace{-.3cm}
\end{figure*}
}
\section{Additional Theoretical Results}
\subsection{Proof of the Optimal Bidding Theorem}\label{sec:optbid}
    
\begin{theorem}
    The optimal bidding function for problem~\eqref{obj} is:
    \begin{equation}
        b_i = \frac{\lambda_0+\lambda_1}{\lambda_1L+\lambda_2}~u_i
    \end{equation}\label{optbfunc}
    where $[\lambda_0,\lambda_1,\lambda_2]$ is a non-zero vector, with $\lambda_i\ge 0,i\in\{0,1,2\}$.
\end{theorem}\label{thm2}
\begin{proof}
Assume the market price $m$ for an impression $\x$ follows a distribution $\mprice$. The expected cost and the expected delivery are as follows:
\begin{equation}
   \begin{aligned}
    &\E[c|\x] = \int_0^{b} c(m)\cdot  \mprice dm\\
    &\E[d|\x] = d \int_0^{b} \mprice dm\\
   \end{aligned} 
\end{equation}
Particularly, in second price auctions, $c_i  \equiv m_i$, and  we write $c_i\equiv c(m_i)$ to indicate that cost is a function of the market price. Note that, $\nabla_b \E [c_i|\x] = b_i\cdot \mpricei$. 

Problem~\eqref{obj} is re-phrased as follows considering the stochasity from $\mprice$:
\begin{equation}
    \begin{aligned}
        \max_{\bb} \quad&h(\bb)\\
    \text{s.t.}\quad & f(\bb)\le 0,~g(\bb)\le0
    \end{aligned}
\end{equation}
where 
\begin{equation}
    \begin{aligned}
        &h(\bb)) \defeq \sum_{i=1}^T \E[d|\x_i]
        ,~f(\bb)\defeq \sum_{i=1}^T -\E[d|\x_i]+L\cdot \E[c|\x_i]\\
        &g(\bb)\defeq \sum_{i=1}^T \E[c|\x_i]-B
    \end{aligned}
\end{equation}

By the Fritz John conditions, the optimal solution satisfies
\begin{equation}
    \lambda_0 \nabla_\bb h -  \lambda_1\nabla_\bb f_j -\lambda_{2} \nabla_\bb g = 0
\end{equation}
where $\lambda = [\lambda_0,\dots,\lambda_2]$ is a non-zero vector such that $\lambda_k\ge 0,\forall k\in\{0,1,2\}$. That means, $\forall i\in\{1,\dots,T\}$
\begin{equation}
    \left(\lambda_0~d_i-\lambda_{2}~b_i+ \lambda_1 (d_i-L b_i)\right)\cdot \mpricei = 0
\end{equation}
which gives the optimal bidding function Eq.~\eqref{optbfunc}, with $u_i = \E[d_i]$.

\end{proof}
\subsection{Derivation of the ELBO objective}\label{sec:elbo}
The objective of Q-learning is to minimize the Bellman Residual:
\begin{equation}
    \E_{(o_t,a_t, o^\prime_t, r_t)\sim\mathcal{B}}\left[ \left(Q(o_t, a_t)-\left(r_t+\gamma \max_a Q(o_{t+1},a)\right)\right)^2\right]\label{br}
\end{equation}
where $\mathcal{B}$ denotes a replay buffer. 

Minimizing the Bellman Residual is equivalent to maximizing the log likelihood of the transition tuple $(o_t,a_t,s_{t+1},r_t)$ with proper assumptions\footnote{The distribution is Gaussian with constant std-dev, and the mean function as $Q(o_t,a_t)$.}:
\begin{equation}
    \begin{aligned}
        &\max \log~P(o_t,a_t,y_t)\\
    = &\max\log~P(y_t|o_t,a_t) P(o_t,a_t)\\
    = &\max - \left(Q(o_t, a_t)-\hat{y}_t\right)^2 + \log P(o_t,a_t)
    \end{aligned}\label{equiv}
\end{equation}
where $\hat{y}_t = r_t+\gamma \max_a Q(o_{t+1},a)$ is the target value computed with the full state $s_{t+1}$. Note that when maximizing the Q function, the second term is treated as constant. 

We use $z$ as the real-valued vector representation for the unobserved state. We have the following \emph{Evidence Lower Bound} for the log likelihood:
\begin{equation}
    \begin{aligned}
        &\log~P(o_t,a_t,y_t)\\
        \ge~&\E_{z\sim q}\left[\log~P(o_t,a_t,y_t|z) \right]- \mathcal{D}_{KL}\left(q(z)\|P(z)\right)\\
        =~&\E_{z\sim q}\left[\log~P(y_t|o_t,a_t,z) \right] + \log~P(o_t,a_t) - \mathcal{D}_{KL}\left(q(z)\|P(z)\right)
    \end{aligned}
\end{equation}

Similar to Eq.~\eqref{equiv}:
\begin{equation}
    \log~P(y_t|o_t,a_t,z)\Leftrightarrow \left(Q(o_t, a_t, z)-y_t\right)^2
\end{equation}

Accordingly, minimizing Eq.~\eqref{br} amounts to maximizing the following ELBO:
\begin{equation}
    \begin{aligned}
        \max_q &~{\E}\left[
                -\E_{z\sim q}\left[
                    \left(Q(o_t, a_t, z)-y_t\right)^2
                \right]
                -\mathcal{D}_{KL}\left(q(z)\|P(z)\right)
            \right]
    \end{aligned}
\end{equation}

\section{Implementation Details}
\GOverallPlot
\OODPlot
\subsection{Curriculum Design}\label{sec:thresh_design}


We have mentioned the design principles for $L^k_t$ in Sec.~\ref{sec:cl}. We implement each curriculum as a dense reward function of the form~\eqref{rdense} with the constraint limits $L_t^k$ evolving along time following the power law:
\begin{equation}
    \begin{aligned}
        &L_t^k = \left(1-b_k\cdot (1-t/T)^g\right)\cdot L\\
        &B_t^k = \left(h_k\cdot (1-t/T)^g\right)\cdot B\label{clhp}
    \end{aligned}
\end{equation}
where $b_k\in [0,1]$ determines the relaxation of the original constraint $L$. For example, $b_k=0.5$ indicates the maximal relaxation of $L$ is by one half, at the beginning of the bidding process. $b_k,h_k$ depends on data, and for fixed curriculum we set $b_k$ to 0.1 and 0.2 for the first two curriculum, and $h_k$ fixed to 0.95. We empirically set $g=3$. To achieve automated curriculum learning, the objective is regret minimization,
\begin{equation}
    \min_{b_k} \E_{\epsilon_T}\left[\left(D_T- \sum_{t=1}^T \R^\prime(s_t,a_t)\right)\1_F\right].
\end{equation}
The insight is to learn $b_k$ such that the proxy cumulative return $\sum_{i=1}^T \R^\prime(s_t,a_t)$ is close to the ground truth $D_T$ given that the episode $\epsilon_T$ is feasible. While reward function~\eqref{rdense} is a function of $b_k$, it is not differentiable due to the indicator function. Accordingly, we use the following smooth approximation for $\1_{F_{L_t^k}}$:
\begin{equation}
    I(\roi_T;v) = \frac{1}{\exp\left(-v\left(x+\sqrt{v}\right)\right)}
\end{equation}
where $v$ controls the slope of the above function transitioning from $0$ to $1$. We empirically set to $10$ with learning rate $3e-3$. 
    


\subsection{Implementation}\label{sec:impl}
\noindent\textbf{Policy design.} The proposed model includes a variational encoder parameterized by a three-layer bi-directional transformer, a conditional policy, a critic that comprises two Q networks and two target Q networks, all implemented as MLPs. For more configurations please refer to our code. 

The input to the policy includes the following statistics: (1) the time slot $t$, (2) the bid ratio in previous slot $b_{t-1}$, 
(3) the current ROI difference $\roi_{t-1}-L$ and the current budget consumption rate $C_{t-1}/B$, 
4) the ROI difference of previous slot $\frac{D_{t-2:t-1}}{C_{t-2:t-1}}-L$, 
5) the normalized delivery of previous slot $T \times D_{t-2:t-1}/D^\ast_T$, 
6) the current surplus $D_{t-1}-L\times C_{t-1}$. Clipping is adopted to ensure the statistics remain in the proper scale. The time slot length is empirically set to half an hour, and thus $T=48$. The output space is set to $[0,4]$ with tanh Gaussian trick to bound the action. 

In \cite{MCB}, temporally correlated action space is used, i.e., policy output is added to the previous bid ratio. Models in our experiments use independent action space except for \textbf{USCB-DDPG}. Besides, we note that \textbf{USCB-DDPG} is different from the standard DDPG~\cite{DDPG} as it fits the Q-function with Monte Carlo return estimates instead of the (bootstrapped) Q-learning. 

\noindent\textbf{Policy Learning.} We adopt SAC~\cite{sac} for policy optimization, an actor-critic method~\cite{DDPG} that uses entropy regularization. We normalize the objective value by the oracle and the constraint violations by the limits, to balance the scale of the two parties. 
The learning rate is set $3e-4$ for all networks, and is stepped at $\{4000,8000,12000\}$ with decay rate $0.5$. 

\noindent\textbf{Slot-wise Oracle.}
We solve the following programming problem,
\begin{equation}
    \max \sum_{t=1}^T D(\epsilon_{[t]};\beta_t), \quad \text{s.t.}~\roi_T\ge L, B-C_t\ge 0,\label{plan2}
\end{equation}
where $D(\epsilon_{[t]};\beta_t,u_{[t]})$ is the total delivery obtained by bidding $\beta_t$ to the impressions in slot $t$ with utilities $u_{[t]}$. 

If we discretize the bid ratio space, the problem~\eqref{plan2} can be treated as a group knapsack problem. Suppose we have $T$ groups of items, in group $t$, an item $\beta_t$ is with value $D(\epsilon_{[t]};\beta_t)=\sum_{i\in[t]} d_i\1_{\beta_t u_i>m_i}$ and weight $C(\epsilon_{[t]};\beta_t)=\sum_{i\in[t]} c_i\1_{\beta_t u_i>m_i}$. Problem~\eqref{plan2} equals to solving the optimal item set such that, (1) in each group, one and only one item is selected, (2) the total weight of selected items does not exceed $B$, and (3) the total value-to-weight ratio should be above $L$. We solve this knapsack problem by an linear programming toolbox\footnote{\href{https://github.com/coin-or/pulp}{https://github.com/coin-or/pulp}} to obtain the slot-wise oracle. This oracle performs better than a day-wise oracle when there are utility prediction error. Intuitively, slot-wise policy lends more space to correct the prediction error by adjusting the bid ratio, while the day-wise policy offers no such flexibility. 

\noindent\textbf{Evaluation Protocols.}
The metrics used in our experiments are computed as:
\begin{align}
    &\text{ANS}\defeq \frac{1}{ N}\sum_{i=1}^N \frac{D(\epsilon_T^{(i)})}{D^\ast(\epsilon_T^{(i)})}\cdot \1_
        {F\left(\epsilon_T^{(i)};L^{(i)},B^{(i)}\right)}\\
    &\text{CSR}\defeq \frac{1}{N}\sum_{i=1}^T \1_{F\left(\epsilon_T^{(i)};L^{(i)},B^{(i)}\right)}\\
    &\text{ANDR}\defeq \frac{1}{|F|}\sum_{\epsilon_T^{(i)}\in F}\left(\frac{D(\epsilon_T^{(i)})}{D^\ast(\epsilon_T^{(i)})}-1\right)\times 100\%
\end{align}


\vspace{-.2cm}
\section{Additional Empirical Results}

The performance of competing methods are shown in Fig.~\ref{fig:oodplot}.

USCB~\cite{MCB} propose a G-metric, which non-linearly combines the constraint violations and the performance objective with a hyper-parameter $\lambda$. 

The main drawback of the G-metric is how to choose the hyper-parameter. We follow \cite{MCB} to use the hyper validated by USCB, i.e., the hyper-parameter with which USCB achieves the best performance in ANS. The results in G-metric are shown in Fig.~\ref{fig:overall_gscore}.

\end{document}